%% file: paper.tex
\documentclass[]{bytedance_seed}

\usepackage[toc,page,header]{appendix}

\input{resources/packages}

\input{resources/math_macro}
\usepackage{mathrsfs}
\usepackage{adjustbox}
\usepackage{multirow}
\usepackage{multirow}
\usepackage{multicol}
\usepackage{tcolorbox}
\usepackage{changepage}
\usepackage{graphicx}
\usepackage{amssymb}
\usepackage{array}

\usepackage{minitoc}

\title{ Seedream 2.0: A Native Chinese-English Bilingual Image Generation Foundation Model }

\author[]{Seed Vision Team, ByteDance}

\abstract{
Rapid advancement of diffusion models has catalyzed remarkable progress in the field of image generation. However, prevalent models such as Flux, SD3.5 and Midjourney, still grapple with issues like model bias, limited text rendering capabilities, and insufficient understanding of Chinese cultural nuances. To address these limitations, we present Seedream 2.0, a native Chinese-English bilingual image generation foundation model that excels across diverse dimensions, which adeptly manages text prompt in both Chinese and English, supporting bilingual image generation and text rendering. 
We develop a powerful data system that facilitates knowledge integration, and a caption system that balances the accuracy and richness for image description. Particularly, Seedream is integrated with a self-developed bilingual large language model (LLM) as a text encode, allowing it to learn native knowledge directly from massive data. This enable it to generate high-fidelity images  with accurate cultural nuances and aesthetic expressions described in either Chinese or English. Beside, Glyph-Aligned ByT5 is applied for flexible character-level text rendering, while a Scaled ROPE generalizes well to untrained resolutions. Multi-phase post-training optimizations, including SFT and RLHF iterations, further improve the overall capability.
Through extensive experimentation, we demonstrate that Seedream 2.0 achieves state-of-the-art performance across multiple aspects, including prompt-following, aesthetics, text rendering, and structural correctness. Furthermore, Seedream 2.0 has been optimized through multiple RLHF iterations to closely align its output with human preferences, as revealed by its outstanding ELO score. In addition, it can be readily adapted to an instruction-based image editing model, such as SeedEdit \cite{seededit2024}, with strong editing capability that balances instruction-following and image consistency.
}

\correspondence{Authors are listed in  \cref{contributions}.}

\checkdata[Official Page]{\url{https://team.doubao.com/tech/seedream} }

\begin{document}
\begin{CJK*}{UTF8}{gbsn}

\maketitle

\definecolor{chinese_red}{HTML}{8B4513}
\definecolor{english_blue}{HTML}{4169E1}

\begin{figure}[ph]
\begin{center}
\vspace{-30pt}
\includegraphics[width=0.90\linewidth]{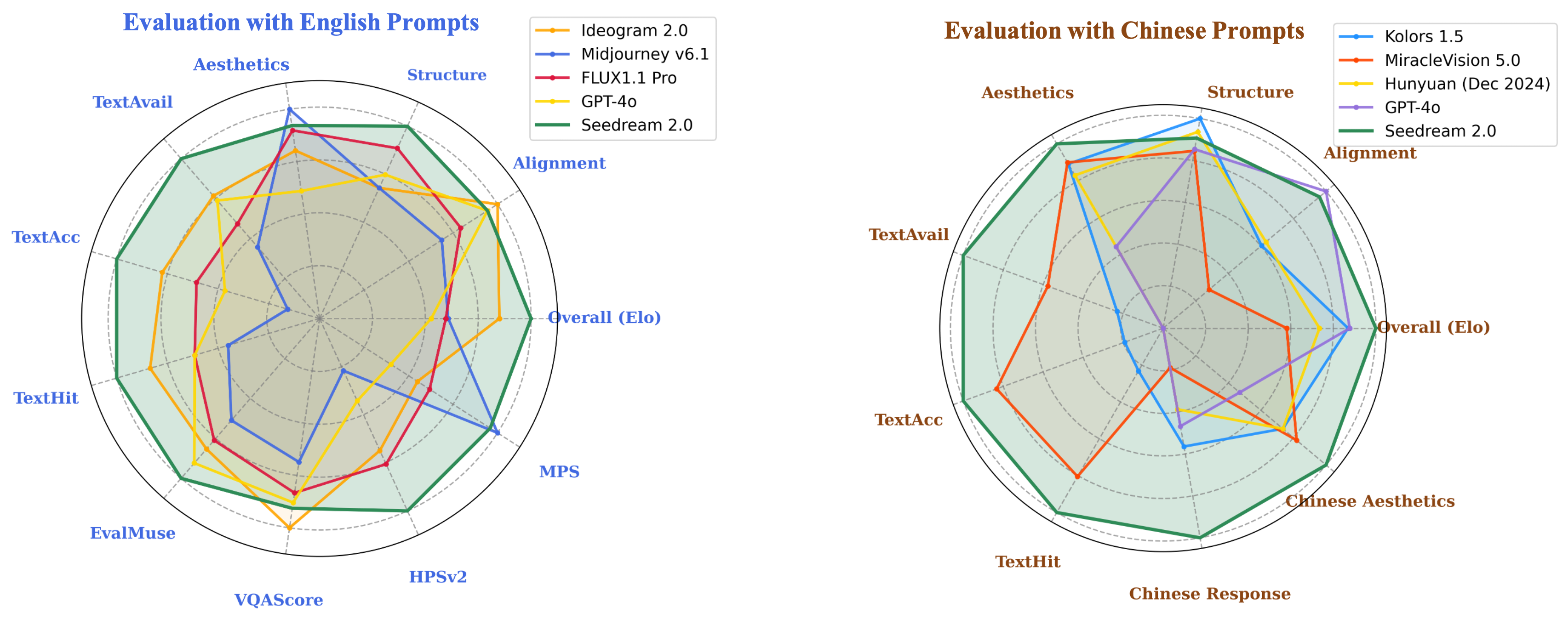}
\end{center}
\label{fig:overall_eval}
\vspace{-5pt}
\caption{Seedream2.0 demonstrates outstanding performance across all evaluation aspects in both \textcolor{english_blue}{English} and \textcolor{chinese_red}{Chinese}.}
\vspace{-8pt}
\end{figure}

\begin{figure}[pt]
\begin{center}
\includegraphics[width=0.9\linewidth]{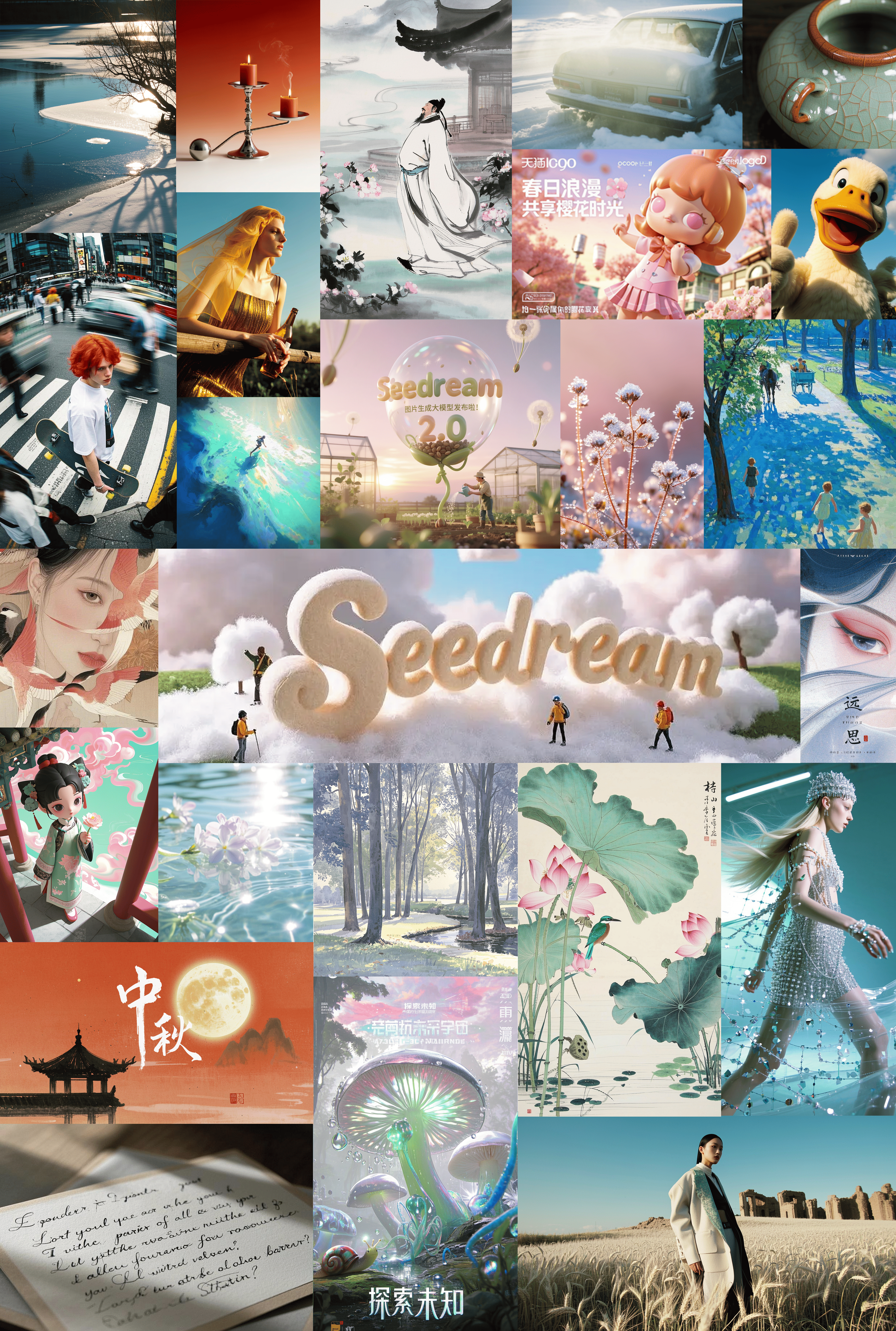}
\end{center}
\label{fig:teaser}
\vspace{-1pt}
\caption{Seedream 2.0 Visualization.}
\end{figure}


\clearpage

\tableofcontents

\newpage

\input{sections/Introduction}


\input{sections/Data_Pre_processing}


\input{sections/Model_Pre_train}


\input{sections/Model_Post_train}


\input{sections/ImageEditing}


\input{sections/Model_Acceleration}

\input{sections/Model_Performance}

\clearpage

\bibliographystyle{plainnat}
\bibliography{main}

\clearpage

\beginappendix
\input{sections/appendix}

\end{CJK*}
\end{document}

%% file: resources/packages.tex
\usepackage{natbib}

\usepackage{CJKutf8}

\usepackage{xargs}  

\usepackage{todonotes}  

\usepackage{multirow}

\usepackage{cleveref}

\usepackage{amsmath}
\usepackage{dsfont}

\usepackage{subcaption}

\usepackage{svg}

%% file: sections/Introduction.tex
\section{Introduction}
With the significant advancement of diffusion models, the field of image generation has experienced rapid expansion. Recent powerful models such as Flux \cite{flux2023}, SD3.5 \cite{esser2024scaling}, Ideogram 2.0, and Midjourney 6.1 have initiated a wave of widespread commercial applications.
However, despite the remarkable progress made by the existing foundational models, they still encounter several challenges.

\begin{itemize}[leftmargin=*]
\item \textbf{\textit{Model Bias}}:
Present models exhibit a propensity towards a specific aspect, such as aesthetics by Midjourney, while sacrificing the performance in other aspects, such as prompt-following or structural correctness.
\item \textbf{\textit{Inadequate Text Rendering Capacity}}:
The ability to perform accurate text rendering in long content or in multiple languages (especially in Chinese) is rather limited, while text rendering is a key ability to some important scenarios, such as a design scenario including graphic design and poster design.
\item \textbf{\textit{Deficiency in Understanding Chinese Characteristics}}:
There is a lack of a deep understanding of the distinctive characteristics of local culture, such as Chinese culture, which is of great importance to local designers. 
\end{itemize}

To address these important issues, we introduce Seedream 2.0, a cutting-edge text-to-image model. It can proficiently handle both Chinese and English prompts, supports bilingual image generation and text rendering tasks, with outstanding performance in multiple aspects.
Specifically, we design a data architecture with the ability to continuously integrate new knowledge,
and develop a strong caption system that considers both \textit{accuracy} and \textit{richness}.
Importantly, we have integrated a self-developed large language model (LLM) as a text encoder with a decoder-only architecture. Through multiple rounds of calibration, the text encoder can obtain enhanced bilingual alignment capabilities, endowment it with native support for learning from original data in both Chinese and English. We also apply a Glyph aligned ByT5 model, which enables our model to flexibly undertake character-level text rendering. Moreover, a Scaled ROPE is proposed to generalize our generation process to untrained  image resolutions.
During a post-training stage, we have further enhanced the model's capabilities through multiple phases of SFT training and RLHF iterations.
Our key contributions are fourfold:

\begin{itemize}[leftmargin=*]
\item \textbf{\textit{Strong Model Capability}}: Through multi-level optimization consisting of data construction, model pre-training, and post-training, our model stands at the forefront across multiple aspects, including prompt-following, aesthetic, text-rendering, and structural correctness.

\item \textbf{\textit{Excellent Text Rendering Proficiency}}:
Using a custom character-level text encoder tailored for text rendering tasks, our model exhibits excellent capabilities for text generation, particularly excelling in the production of long textual content with complicated Chinese characters.

\item \textbf{\textit{Profound Understanding of Chinese Characteristics}}: By integrating with a self-developed multi-language LLM text encoder, our model can learn directly from massive high-quality date in Chinese. This makes it powerful to handle complicated prompts infused with Chinese stylistic nuances and specialized professional vocabulary. Furthermore, our model demonstrates exceptional performance in Chinese text rendering, which is not well developed in the community.

\item \textbf{\textit{Highly Align with Human Preferences}}: Following multiple iterations of RLHF optimizations across various post-training modules, our model consistently aligns its outputs with human preferences, which is evidenced by a great advantage in ELO scoring.

\end{itemize}

As of early December 2024, Seedream2.0 has been incorporated into various platforms exemplified by Doubao (豆包) \protect\footnotemark[1] and Dreamina (即梦) \protect\footnotemark[2]. We ardently encourage a broader audience to delve into the extensive capabilities and potentials of our model, with the aspiration that it can emerge as an effective tool for improving productivity in the multiple aspects of work and daily life.

\footnotetext[1]{https://www.doubao.com/chat/create-image}
\footnotetext[2]{https://jimeng.jianying.com/ai-tool/image/generate}

%% file: sections/Data_Pre_processing.tex
\section{Data Pre-Processing}
This section details our data pipeline for pre-training, encompassing various pre-processing steps such as data composition, data cleaning and filtering, active learning, captioning, and data for text rendering. These processes ensure a final pre-training dataset that is of high quality, large scale, and diverse.

\subsection{Data Composition}
\begin{figure*}[h]
\centering
\includegraphics[width=0.6\linewidth]{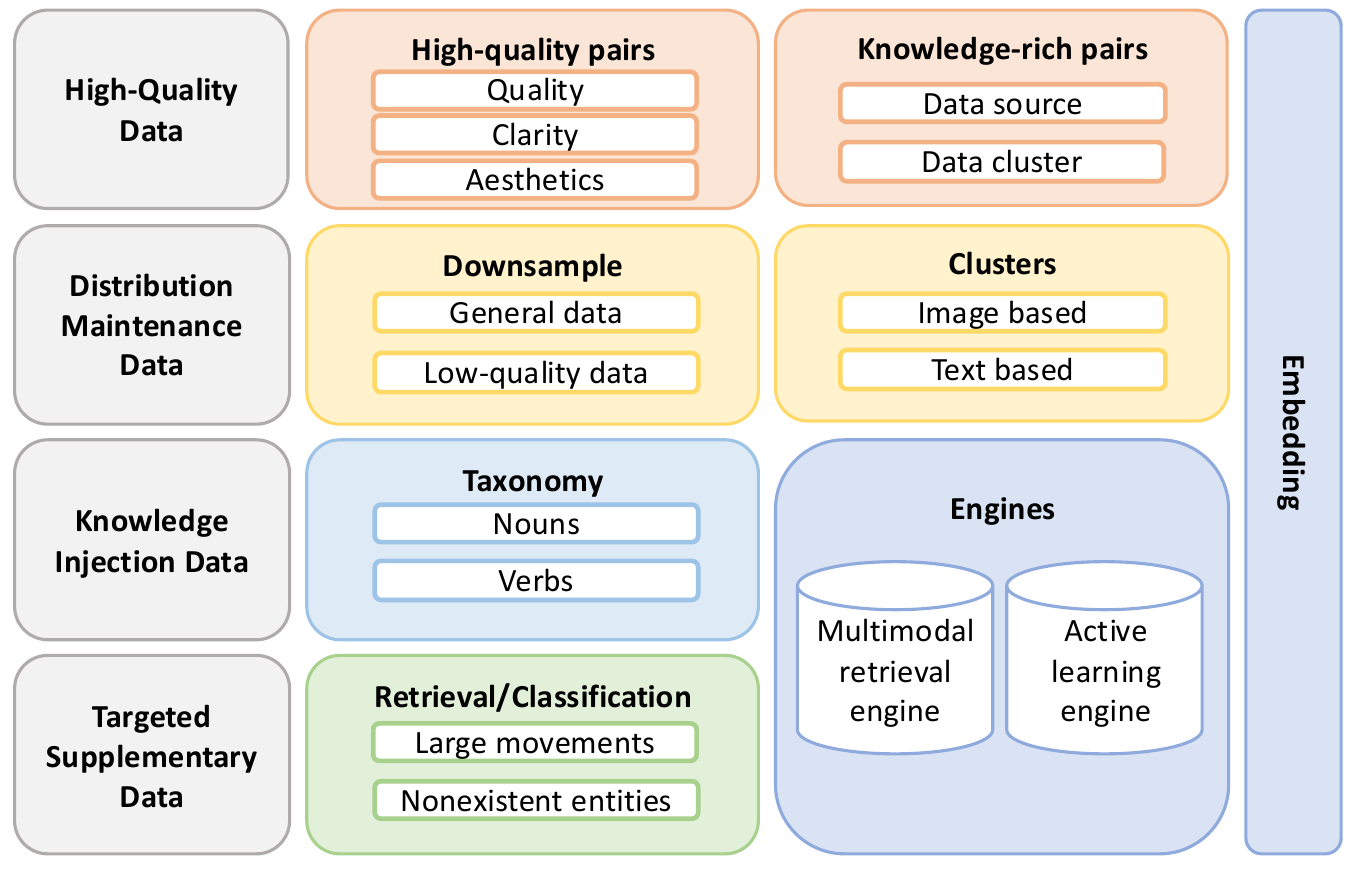}
\caption{Pre-training data system.}
\label{fig:data}
\end{figure*}

Our pre-training data is meticulously curated from four main components, ensuring a balanced and comprehensive dataset, as shown in Figure \ref{fig:data}.

\textbf{High-Quality Data.} This component includes data with exceptionally high image quality and rich knowledge content, assessed based on clarity, aesthetic appeal, and source distribution.

\begin{figure*}[h]
\centering
\includegraphics[width=0.6\linewidth]{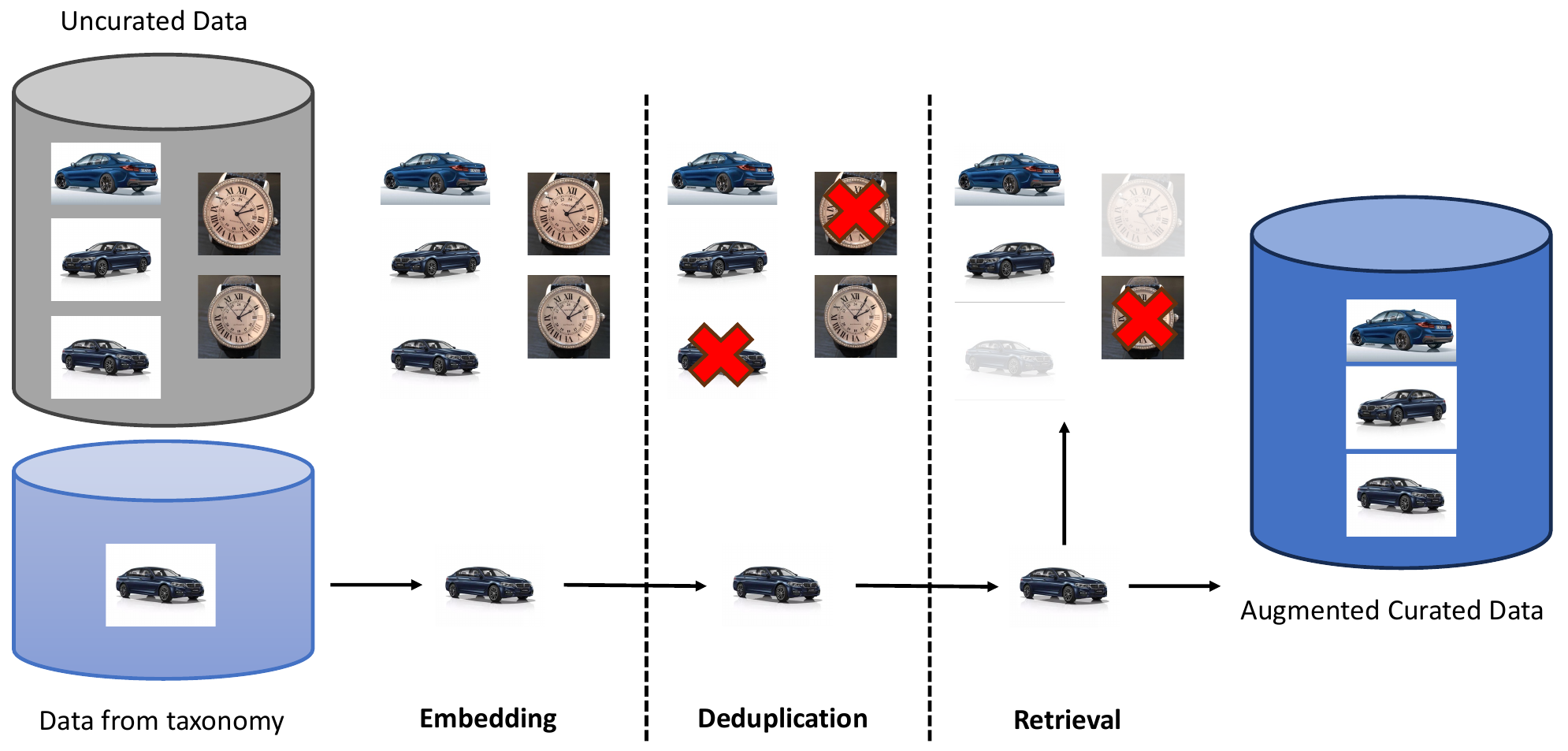}
\caption{Overview of our knowledge injection process.}
\label{fig:knowledge_injection_data}
\end{figure*}

\textbf{Distribution Maintenance Data.} This component maintains the useful distribution of the original data while reducing low-quality data through:
\begin{itemize}
\item \textbf{Downsampling by Data Source:} Reducing the proportion of overrepresented sources while preserving their relative magnitude relationships.
\item \textbf{Clustering-based Sampling:} Sampling data based on clusters at multiple hierarchical levels, from clusters representing broader semantics (such as visual designs) to those representing finer semantics, e.g., CD/book covers and posters.
\end{itemize}

\textbf{Knowledge Injection Data.} This segment involves the injection of knowledge using a developed taxonomy and a multimodal retrieval engine, as shown in Figure \ref{fig:knowledge_injection_data}. It includes data with distinctive Chinese contexts to improve model performance in Chinese-specific scenarios.

Additionally, a small batch of data with distinctive Chinese contexts was manually collected. This dataset includes image-text pairs related to Chinese-specific characters, flora and fauna, cuisine, scenes, architecture, and folk culture.
Our multimodal retrieval engine was employed to augment and incorporate this Chinese knowledge into our generative model.

\textbf{Targeted Supplementary Data.} We supplement the dataset with data that exhibit suboptimal performance in text-to-image tasks, such as action-oriented data and counterfactual data (e.g., "a man with a balloon for a neck"). Our active learning engine categorizes and integrates these challenging data points into the final training set.

\subsection{Data Cleaning Process}
\begin{figure*}[h]
\centering
\includegraphics[width=0.6\linewidth]{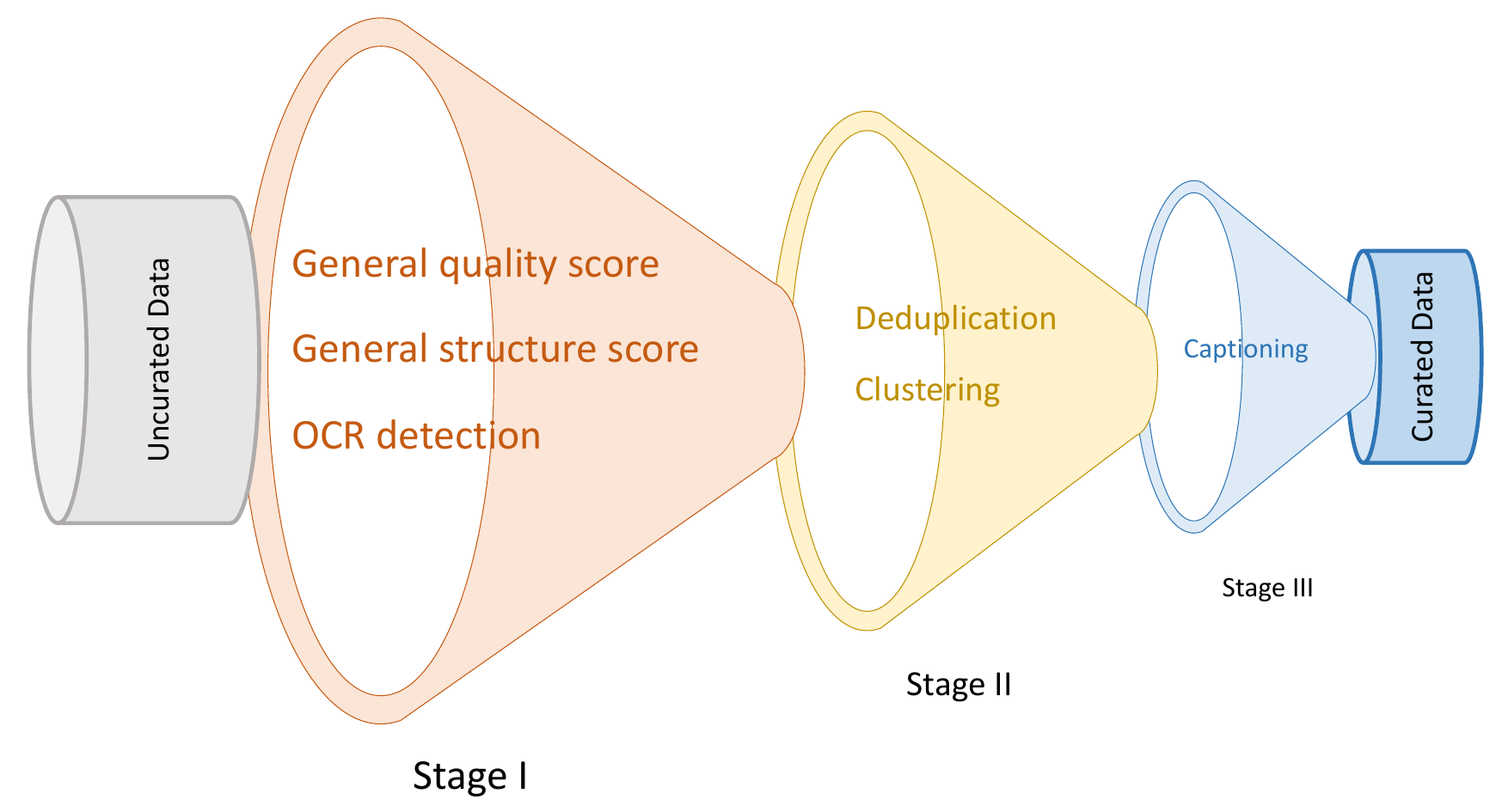}
\caption{Overview of our data cleaning process.}
\label{fig:data_cleaning}
\end{figure*}

The data cleaning procedure ensures the quality and relevance of the dataset through progressively elaborate data filtering methodologies, as depicted in Figure \ref{fig:data_cleaning}.

\textbf{First Stage: General Quality Assessment.} We label the entire database using the following criteria:
\begin{itemize}
\item \textbf{General Quality Score:} Evaluating image clarity, motion blur, and meaningless content.
\item \textbf{General Structure Score:} Assessment of elements such as watermarks, text overlays, stickers, and logos.
\item \textbf{OCR Detection:} Identifying and cataloging text within images.
\end{itemize}
Samples that do not meet quality standards are eliminated.

\textbf{Second Stage: Detailed Quality Assessment.} This stage involves professional aesthetic scores, feature embedding extraction, deduplication, and clustering. Clustering is structured at multiple hierarchical levels, representing distinct semantic categories. Each data point is assigned a semantic category tag for subsequent adjustment of the distribution.

\textbf{Third Stage: Captioning and Re-captioning.} We stratify the remaining data and annotate captions or recaptions. Higher-level data generally receive richer new captions, described from different perspectives. Details on the captioning process are provided in Section \ref{subsection:captioning}.

\subsection{Active Learning Engine}
\begin{figure*}[h]
\centering
\includegraphics[width=0.6\linewidth]{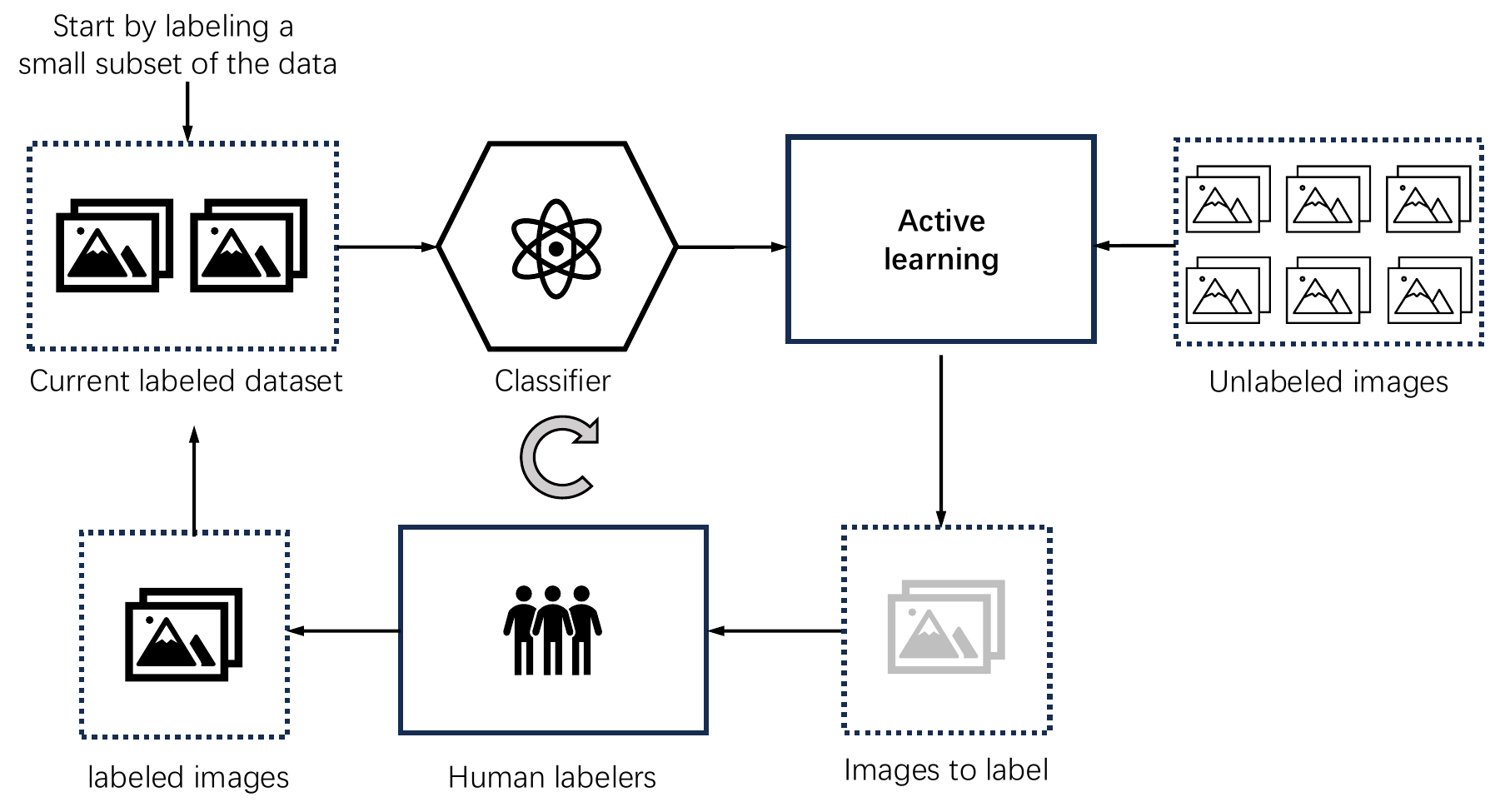}
\caption{Flow diagram of Active Learning Lifecycle.}
\label{fig:active_learning}
\end{figure*}

We developed an active learning system to improve our image classifiers, as illustrated in Figure \ref{fig:active_learning}. It is an iterative procedure that progressively refines our classifiers, ensuring a high-quality dataset for training.

\subsection{Image Captioning} \label{subsection:captioning}
The captioning process provides meaningful and contextually accurate descriptions for each image, generating both generic and specialized captions.

\subsubsection{Generic Captions}
We formulate short and long captions in Chinese and English, ensuring accurate and detailed descriptions:
\begin{itemize}
\item \textbf{Short Captions:} Accurately describe the main content of an image, capturing the core knowledge and content.
\item \textbf{Long Captions:} More descriptive, detailing as many aspects of an image as possible, including appropriate inferences and imaginations.
\end{itemize}

\begin{figure*}[h]
\centering
\includegraphics[width=\linewidth]{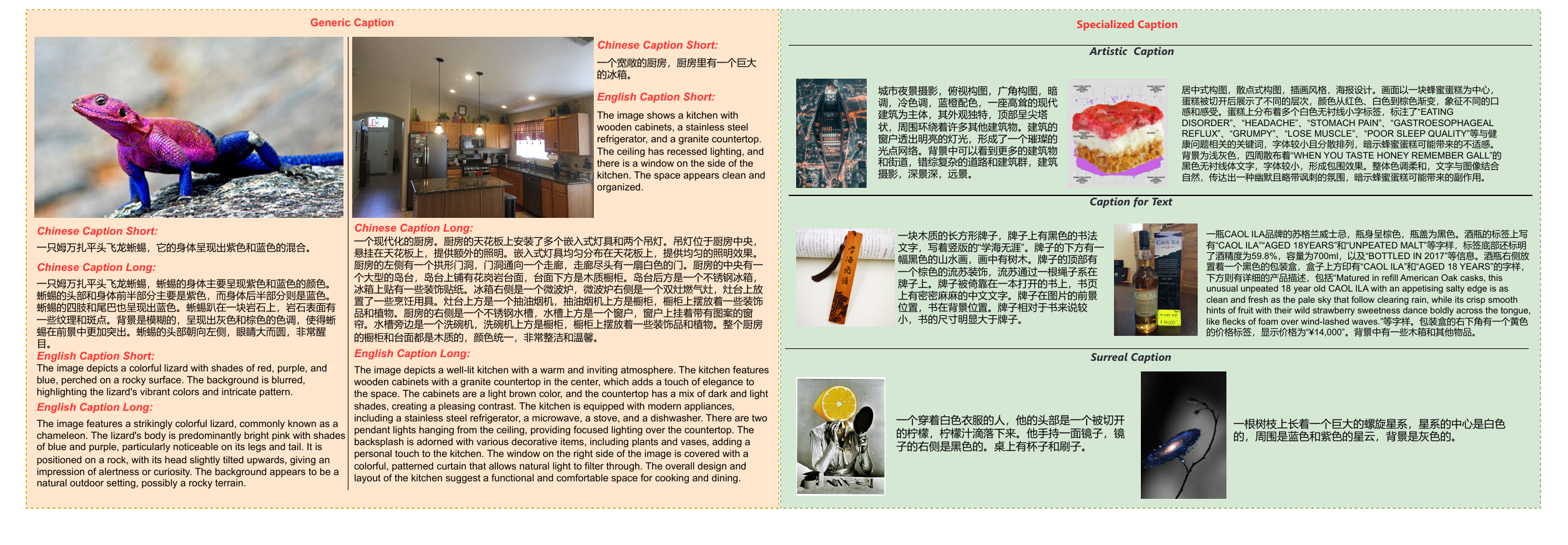}
\caption{Caption examples in our training data.}
\label{fig:caption}
\end{figure*}

\subsubsection{Specialized Captions}
In addition to generic captions, we also have specialized captions designed for various distinct scenarios:
\begin{itemize}
\item \textbf{Artistic Captions:} Describe aesthetic elements such as style, color, composition, and light interaction.
\item \textbf{Textual Captions:} Focus on the textual information present in the images.
\item \textbf{Surreal Captions:} Capture the surreal and fantastical aspects of images, offering a more imaginative perspective.
\end{itemize}

\begin{figure*}[th]
\centering
\includegraphics[width=0.8\linewidth]{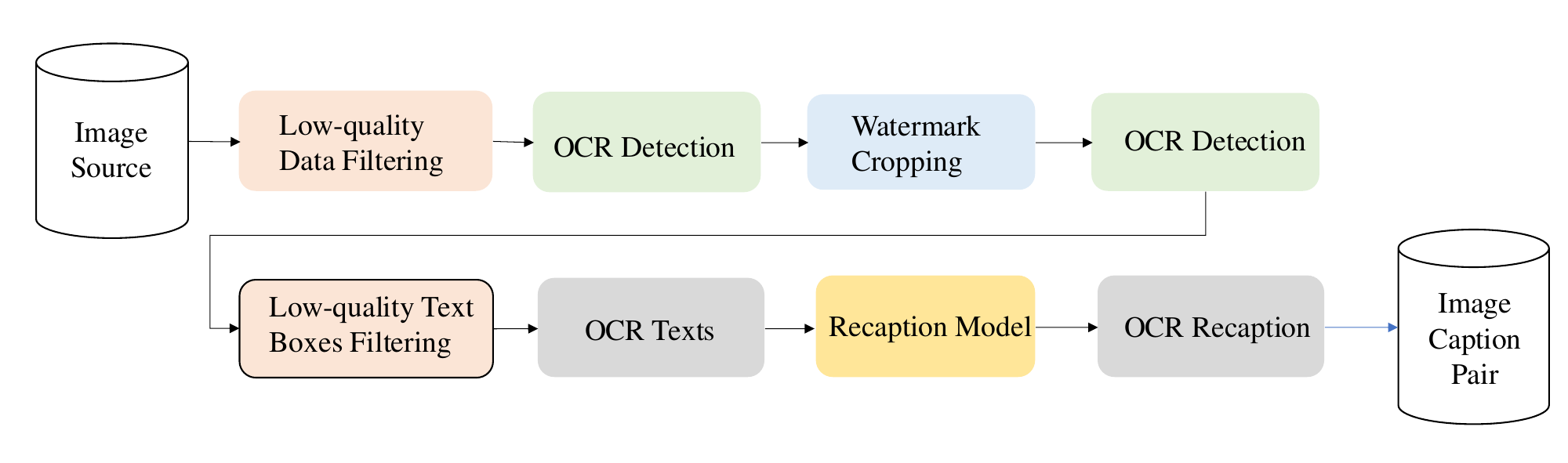}
\caption{Text Rendering: Data Pre-processing Pipeline.}
\label{fig:visual_text_pipeline}
\end{figure*}

\subsection{Text Rendering Data}
We construct a large-scale visual text rendering dataset by filtering in-house data and using OCR tools to select images with rich visual text content, as depicted in Figure \ref{fig:visual_text_pipeline}. The main data processing steps are as follows:
\begin{itemize}
\item Filter low-quality data from in-house sources.
\item Employ OCR to detect and extract text regions, followed by cropping of watermarks.
\item Remove low-quality text boxes, retaining clear and relevant text regions.
\item Process extracted text using a re-caption model to generate high-quality descriptions.
\item Further refine the descriptions to produce high-quality image-caption pairs which are finally used for visual text-rendering tasks.
\end{itemize}

%% file: sections/Model_Pre_train.tex
\section{Model Pre-Training}

\begin{figure*}[h]
\centering
\includegraphics[width=1.0\linewidth]{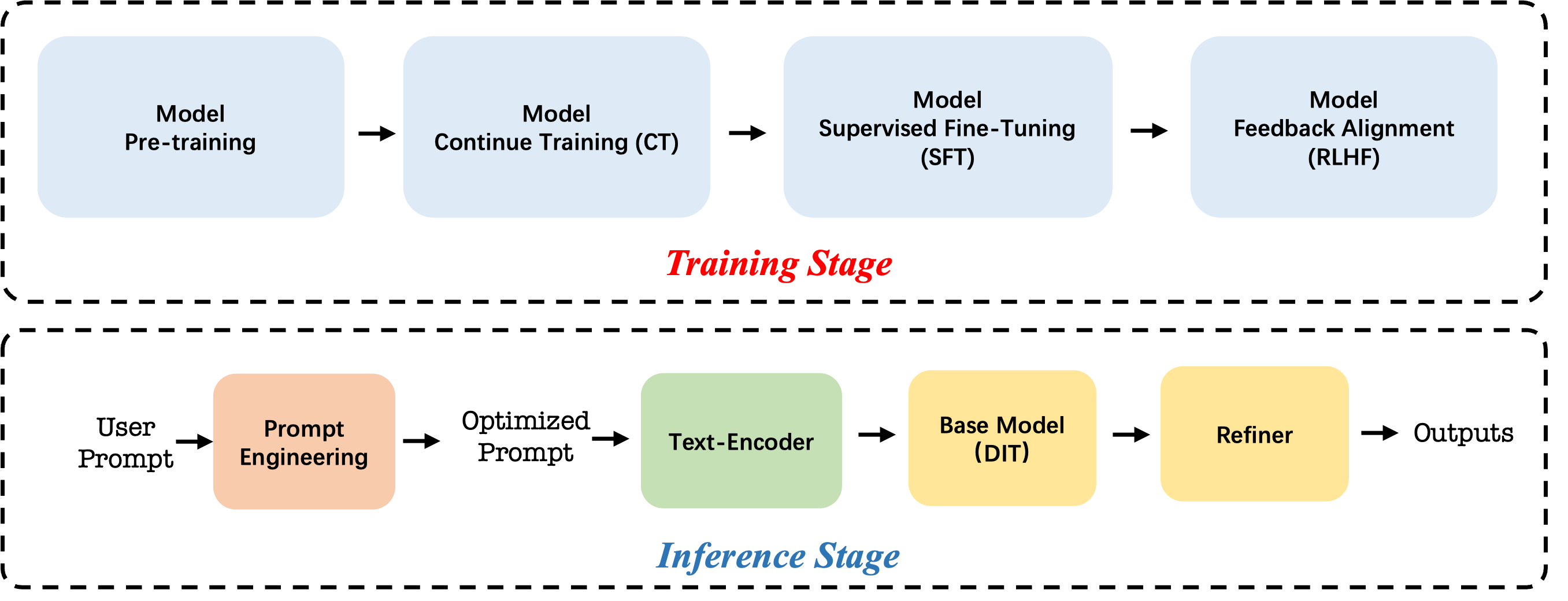}
\caption{Overview of Seedream 2.0 Training and Inference Pipeline.}
\label{fig:Pipeline}
\end{figure*}

In this section, we introduce the training and inference stages of our Seedream 2.0 model. The main modules are presented in Figure \ref{fig:Pipeline}. 

\subsection{Diffusion Transformer}

\begin{figure*}[h]
\centering
\includegraphics[width=1.0\linewidth]{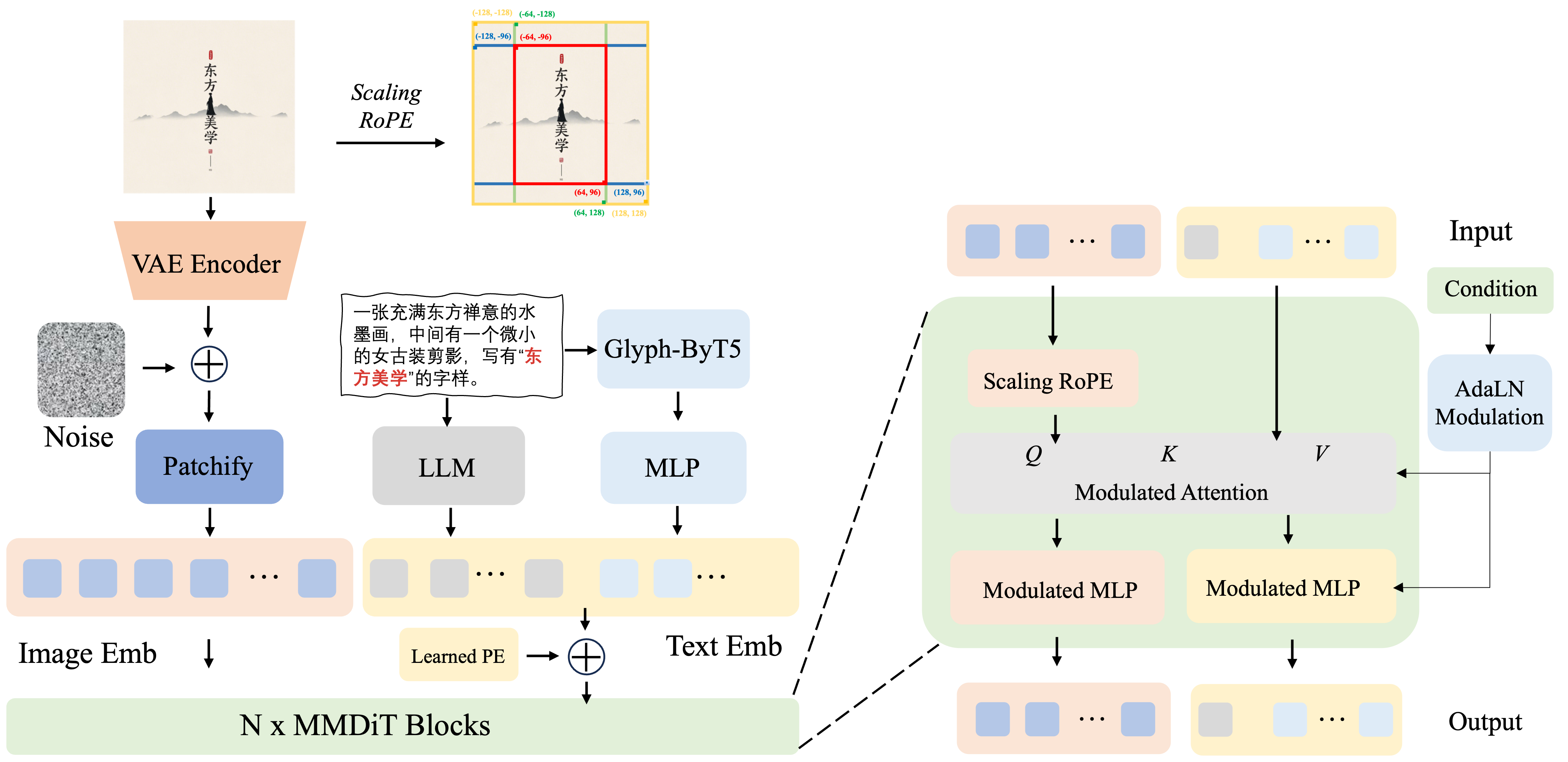}
\caption{Overview of Model Architecture.}
\label{fig:architecture}
\end{figure*}

For an input image \textbf{\textit{I}}, a self-developed Variational Auto-Encoder (VAE)  is used to encode the input image, resulting in a latent space representation $\mathbf{x} \in \mathbb{R}^{C\times H \times W}$. The latent vector $\mathbf{x}$ is then patchified to a number of patches being $\frac{H}{p} \times \frac{W}{p}$. This process ultimately transforms the input image into $\frac{H \times W}{4}$ image tokens, which are concatenated with text tokens encoded by a text encoder and then fed into transformer blocks.

The design of DiT blocks mainly adheres to the design principles of MMDiT in Stable Diffusion 3 (SD3) \cite{esser2024scaling}. Each transformer block incorporates only a single self-attention layer, which concurrently processes both image and text tokens. Considering the disparities between the image and text modalities, distinct MLPs are employed to handle them separately. The adaptive layer norm is utilized to modulate each attention and MLP layer. 
We resort to QK-Norm to improve training stability and Fully Sharded Data Parallel (FSDP) \cite{zhao2023pytorch} to conduct distributed model training. 

In this paper, we add the learned positional embedding on text tokens, and apply a 2D Rotary Positional Embedding (RoPE) \cite{su2024roformer} on image tokens. Unlike previous works, we develop a variant of 2D RoPE, namely Scaling RoPE. As shown in Fig. \ref{fig:architecture}, by configuring various scale factors based on image resolution, the patches located at the center of the image can share similar position IDs across different resolutions. This enables our model to be generalized to untrained aspect ratios and resolutions to a certain extent during inference.

\subsection{Text Encoder}

To perform effective prompt encoding for text-to-image generation models, existing methodologies (\cite{flux2023,li2024hunyuan, esser2024scaling}) typically resort to employing CLIP or T5 as a text encoder for diffusion models. 
CLIP text encoder (\cite{radford2021learning}) is capable of capturing discriminative information that is well aligned with visual representation or embeddings, while the T5 encoder (\cite{raffel2020exploring}) has a strong ability to understand complicated and fine-grained text information. However, neither CLIP or T5 encoder has strong ability to understand text in Chinese, while decoder-only LLMs often have excellent multi-language capabilities.

A text encoder plays a key role in diffusion models, particularly for the performance of text-alignment in image generation. Therefore, we aim to develop a strong text encoder by taking advantage of the power of LLMs that is stronger than that of CLIP or T5. However, text embeddings generated by the decoder-only LLMs have large differences in feature distribution compared to the text encoder of CLIP or T5, making it difficult to align well with image representations in diffusion models. This results in significant instability when training a diffusion model with such an LLM-based text encoder. We develop a new approach to fine-tune a decoder-only LLM by using text-image pair data. To further enhance the capabilities for generating images in certain challenging scenarios, such as those involving Chinese stylistic nuances and specialized professional vocabulary, we
collect a large amount of such types of data included in our training set.

Using the strong capabilities of LLM, and implementing the meticulously crafted training strategies, our text encoder has demonstrated a superior performance over other models across multiple perspectives, including strong bilingual capabilities that enable excellent performance in long-text understanding and complicated instruction following. In particular, excellent bilingual ability makes our models able to learn meaningful native knowledge directly from massive date in both Chinese and English, which is the key for our model to generate images with accurate cultural nuances and aesthetic expressions described in both Chinese and English.

\subsection{Character-level Text Encoder}

Considering the complexity of bilingual text glyphs (especially Chinese characters), we apply a ByT5 \cite{xue2022byt5,liu2024glyph} glyph-aligned model to encode the glyph content of rendered text. This model can provide accurate character-level features or embeddings and ensure the consistency of glyph features of rendered text with that of a text prompt, which are concatenated and then are input into a DIT block.

\textbf{Rendering Content.} Experimental results have demonstrated that when using a ByT5 model solely to encode the features of a rendered text, particularly in the case of long text, it can lead to repeated characters and disordered layout generation. This is due to the model's insufficient understanding of holistic semantics.
To address this issue, for the glyph features of rendered text, we encode them using both an LLM (text encoder) and a ByT5 model. Then we employ an MLP layer to project the ByT5 embeddings into a space that aligns with the features of the LLM text encoder. Then, after splicing the LLM and ByT5 features, we send the complete text features to the DiT blocks for training.
In contrast to other approaches that typically use both LLM features and OCR-rendered image features as conditions, our approach uses only textual features as conditions. This allows our model to maintain the same training and reasoning process as the original text-to-image generation, significantly reducing the complexity of the training and reasoning pipeline.

\textbf{Rendering Features.}
The font, color, size, position and other characteristics of the rendered text are described using a re-caption model which is encoded through an LLM text encoder.
Traditional text rendering approaches \cite{tuo2023anytext,liu2024glyphv1,chen2024textdiffuser} typically rely on a layout of preset text boxes as a conditional input to a diffusion model. For example, TextDiffuser-2 \cite{chen2024textdiffuser} employs an additional LLM for layout planning and encoding. In contrast, our approach directly describes the rendering features of the text through the re-caption model, allowing for an end-to-end training.
This enables our model to learn the rendering features of text effectively and directly from training data, which also makes it efficient to learn accurate glyph features of the rendered text based on the encoded rendering features. This approach allows for a more comprehensive and accurate understanding of the rendering text, enabling the creation of more sophisticated and high-quality text rendering outputs.

%% file: sections/Model_Post_train.tex
\section{Model Post-Training}

\begin{figure*}[p]
\centering
\includegraphics[width=0.98\linewidth]{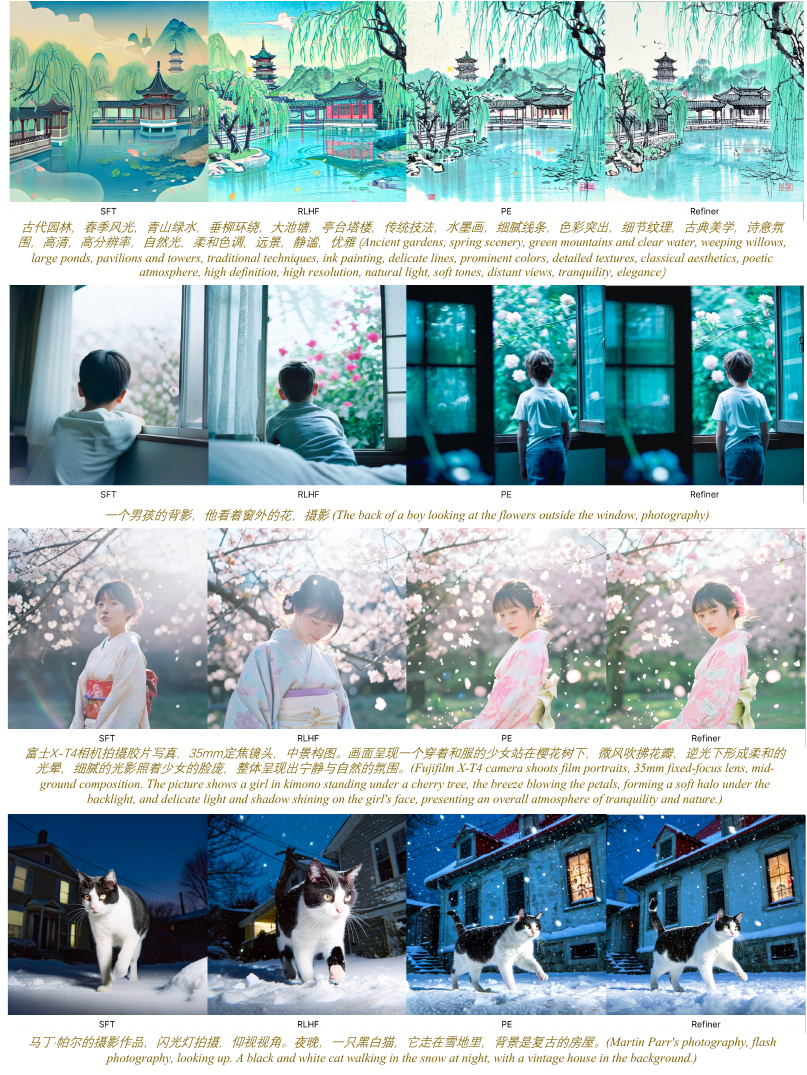}
\caption{Visualization during different post-training stages.}
\label{fig:post}
\end{figure*}

Our post-training process consists of multiple sequential phases: 1) Continue Training (CT) and Supervised fine-tuning (SFT)  stages remarkably enhance the aesthetic appeal of the model; 2) Human Feedback Alignment (RLHF) stage significantly improves the model's overall performance across all aspects via self-developed reward models and feedback learning algorithms; 3) Prompt Engineering (PE) further improves the performance on aesthetics and diversity by leveraging a fine-tuned LLM; 4) Finally, a refiner model is developed to scale up the resolution of an output image generated from our base model, and at the same time fix some minor structural errors. The visualization results during different post-training stages are presented in Figure \ref{fig:post}.

\subsection{Continuing Training (CT)}
Pre-trained diffusion models often struggle to produce images that meet the desired aesthetic criteria, due to the disparate aesthetic standards inherent in the pre-training datasets. To confront this challenge, we extend the training phase by transitioning to a smaller but better quality data set. This continuing training (CT) phase is designed not only to markedly enhance aesthetics of the generated images, but is also required to maintain fundamental performance on prompt-following and structural accuracy. The data of the CT stage consists of two parts.

\subsubsection{Data}

\begin{itemize}[leftmargin=*]

\item ~\underline{\textit{High-quality Pre-training Data}}: We filter a large amount of high-quality images from our pre-training dataset, by developing a series of specialized Image Quality Assessment (IQA) models. The filtering process is automatic by using these models without any manual effort.

\item ~\underline{\textit{Manually Curated Data}}: In addition to the collected high-quality data from pre-training datasets, we meticulously amass datasets with elevated aesthetic qualities from diverse specific domains such as art, photography, and design. The images within these datasets are required to possess a certain aesthetic charm and align with the anticipated image generation outcomes. Following multiple rounds of refinement, a refined dataset comprising millions of manually cherry-picked images was fabricated. To avoid overfitting such a small dataset,  we continually train our model by jointly using it with the selected high-quality pre-trained data, with a reasonable sampling ratio.

\end{itemize}

\subsubsection{Training Strategy}

Directly performing CT on the aforementioned datasets can considerably improve the performance in terms of aesthetics, but the generated images still exhibit a notable disparity from real images having appealing aesthetics. 
To further improve aesthetic performance, we introduce VMix (\cite{wu2024vmix}) which enables our model to learn the \textit{ fine-grained aesthetic characteristics} directly during the denoising process. We tag each image according to various aesthetic dimensions, namely \textit{color}, \textit{lighting}, \textit{texture}, and \textit{composition}, and then these tags are used as supplementary conditions during our CT training process. Experimental results show that our method can further enhance the aesthetic appeal of the generated images.

\subsection{Supervised Fine-Tuning (SFT)}
\subsubsection{Data}
In the SFT stage, we further fine-tune our model toward generating high-fidelity images with excellent \textit{artistic beauty}, by using a small amount of carefully collected images.
With these collected images, we specifically trained a caption model capable of precisely describing beauty and artistry through multi-round manual rectifications.
Furthermore, we also assigned style labels and fine-grained aesthetic labels (used in the vmix approach) to these images, which ensure that the information of the majority of mainstream genres is included.

\subsubsection{Training Strategy}
In addtion to the constructed SFT data, we also include a certain amount of model-generated images, which are labeled as "negative samples", during SFT training. 
By combining with real image samples, the model can learn to discriminate between real and fake images, enabling it to generate more natural and realistic images. This thereby enhances the quality and authenticity of the generated images. 
The SFT data with high artistic standards can substantially enhance the artistic beauty, but it inevitably degrades the performance on image-text alignment, which is fundamental to text-to-image generation task. To address this issue, we developed a data resampling algorithm that allows the model to enhance aesthetics while still maintaining image-text alignment capacity.

\subsection{Human Feedback Alignment (RLHF)}

\begin{figure*}[t]
\centering
\includegraphics[width=\linewidth]{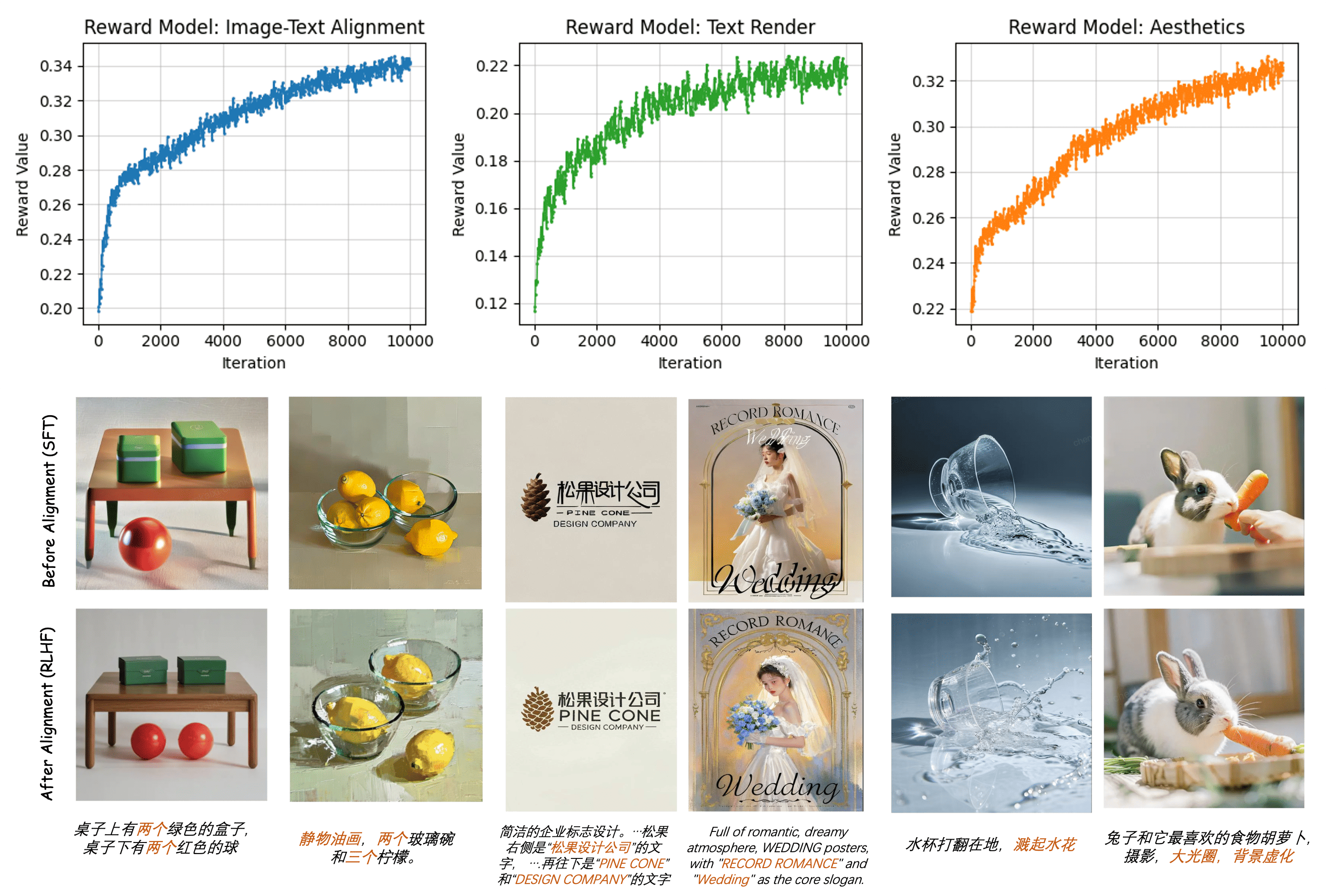}

\caption{The reward curves show that the values across diverse reward models all exhibit a stable and consistent upward trend throughout the alignment process. Some visualization examples reveal that the human feedback alignment stage is crucial.}
\label{fig:Reward}
\end{figure*}

In our work, we introduce a pioneering RLHF optimization procedure tailored for diffusion models (\cite{zhang2024unifl,li2025controlnet,zhang2024onlinevpo}), incorporating preference data, reward models(RMs), and a feedback learning algorithm. As depicted in Figure \ref{fig:Reward}, the RLHF phase plays a pivotal role in enhancing the overall performance of our diffusion models in various aspects, including image-text alignment, aesthetic, structure correctness, text rendering, etc.

\subsubsection{Preference Data}

\begin{itemize}[leftmargin=*]
\item ~\underline{\textit{Prompt System}}: We have developed a versatile Prompt System tailored for employment in both the RM Training and Feedback Learning phases. Our curated collection comprises of 1 million multi-dimensional prompts sourced from training captions and user input. Through rigorous curation processes that filter out ambiguous or vague expressions, we guarantee a prompt system that is not only comprehensive but also rich in diversity and depth of content.

\item ~\underline{\textit{RM Data Collection}}:
We collect high-quality data for preference annotation, comprising images crafted by various trained models and data sources. Through the construction of a cross-version and cross-model annotation pipeline, we enhance the domain adaptability of RMs, and extend its upper threshold of preferences.

\item ~\underline{\textit{Annotation Rules}}: In the annotation phase, we engage in multi-dimensional fusion annotation (such as image and text matching, text rendering, aesthetic, etc.). These integrated annotation procedures are designed to elevate the multi-dimensional capabilities of a single reward model, forestall deficiencies in the RLHF stage, and foster the achievement of Pareto optimality across all dimensions within RLHF.
\end{itemize}

\subsubsection{Reward Model}

\begin{itemize}[leftmargin=*]
\item ~\underline{\textit{Model Architecture}}: We use a CLIP model that supports both Chinese and English as our RMs. By leveraging the strong alignment capabilities of the CLIP model, we forgo additional Head output reward methods like ImageReward, opting to utilize the output of CLIP model as the reward itself.  A ranking loss is primarily applied as the training loss of our RMs. 

\item ~\underline{\textit{Multi-aspects Reward Models}}: To enhance the overall performance of our models, we meticulously crafted and trained three distinct RMs: a image-text alignment RM, an aesthetic RM, and a text-rendering RM. In particular, the text-rendering RM is selectively engaged when a prompt tag relates to text rendering, significantly improving the precision of character-level text generation.

\end{itemize}

\subsubsection{Feedback Learning}

\begin{itemize}[leftmargin=*]
\item ~\underline{\textit{Learning Algorithm}}: We refine our diffusion model through a direct optimization of output scores computed from multiple RMs, akin to REFL (\cite{xu2024imagereward}) paradigm. Delving into various feedback learning algorithms such as DPO (\cite{wallace2024diffusion}) and DDPO (\cite{black2023training}), our investigation revealed that our methods stand out as an efficient and effective approach toward multi-reward optimization. 
In particular, we achieve stable feedback learning training by carefully adjusting learning rates, choosing an appropriate denoising time step, and implementing weight exponential moving average. 
During the feedback learning phase, a pivotal strategy involves harmonized fine-tuning of the DIT and the integrated LLM text encoder. This joint training protocol significantly amplifies the model's capacity in image-text alignment and aesthetic improvement.

\item ~\underline{\textit{Iterative Refinement}}: Our experimentation involves a series of iterative optimizations performed between the diffusion model and the trained RMs. 
i) We begin by utilizing the existing reward model to optimize the diffusion model. ii) Next, we conduct preference annotation on the refined diffusion model and train a bad-case-aware reward model. iii) We then leverage this updated reward model to further optimize the diffusion model. The above process is iteratively repeatting to enhance performance.
This iterative approach not only enhances the upper bound of performance within the RLHF process but also ensures a higher degree of stability and control compared to dynamically updating the RMs.

\end{itemize}

\subsection{Prompt Engineering (PE)}

\begin{figure*}[t]
\centering
\includegraphics[width=\linewidth]{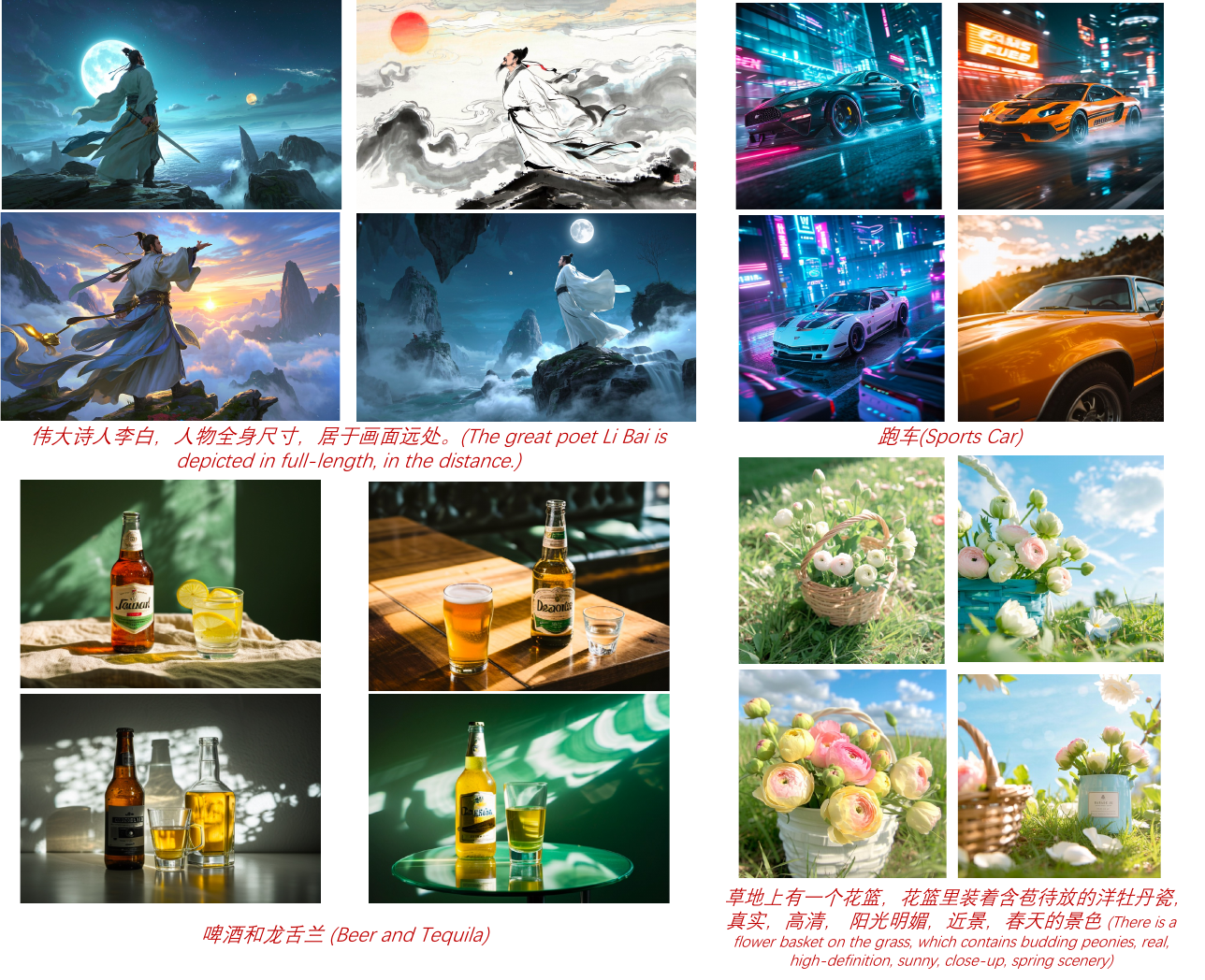}
\caption{PE Visualization. We provide 4 PE prompts for each original prompt.}
\label{fig:pe}
\end{figure*}
A common user text prompt is often simple and uncluttered, and it is difficult to directly generate an image with a satisfied quality.
%
This limitation stems from the fact that our diffusion model is trained with high-quality captions, which are often much more complicated but include more detailed information than human-written text prompt. This means that we need to recalibrate user prompts to match the model's preferences for achieving optimal performance.
To address this issue, we introduce a novel Prompt Engineering (PE) framework, by leveraging an internal finetuned LLM to facilitate diffusion model to generate images with higher quality. The PE framework consists of two key stage: supervised fine-tuning LLM and RLHF. Our empirical findings demonstrate that our PE model leads to a notable 30\% enhancement in the aesthetic quality, a 5\% improvement in image-text alignment, and a substantial increase in diversity of the generated images.

\subsubsection{Fine-tune LLM}
Our PE model is built on a well-developed LLM with strong ability in both Chinese and English. We perform supervised fine-tuning on the LLM, by using a carefully curated dateset, where we construct data samples of paired prompts, $D={<u,r>}$ ($u$ denotes an initial input prompt and $r$ represents a rephrased one output by our PE model).
The quality of the constructed prompt pairs is important for the performance of PE. We devised two distinct methodologies: i) starting from user input: ($u \rightarrow r$): a user input prompt $u$ is manually rephrased and then input into a well-developed T2I diffusion model. This process is implemented repeatedly until a high-quality image is generated, where the corresponding rephrased prompt is selected as $r$.
ii) starting from a rephrased prompt ($r \rightarrow u$): we carefully curate excellent image samples with detailed and comprehensive captions from our training set. Furthermore, we collect such high-quality samples or image-text pairs from the open-source community. Then we degrade the captions of the collected samples to obtain the initial user prompt $u$, by using an internal LLM (for example, to eliminate aesthetic-related descriptions in the rephrased captions).

\subsubsection{PE RLHF}
We perform RLHF on our PE LLM via diffusion generation, which can further enhance the PE model that improves our image generation results with higher aesthetic quality and more accurate image-text alignment. 
Specifically, we collect a set of user prompts from our training data. Then we perform the current PE model on each user prompt to generate multiple rephrased prompts, which are then used to generate images using the trained diffusion model. We select a <high-quality, low-quality> image pair from the generated images, based on aesthetics and text-image alignment results. Finally, the corresponding prompt pairs are used to further train PE using a simple preference optimization (SimPO) method, which further aligns the PE performance to human preferences.

\begin{figure*}[t]
\centering
\includegraphics[width=\linewidth]{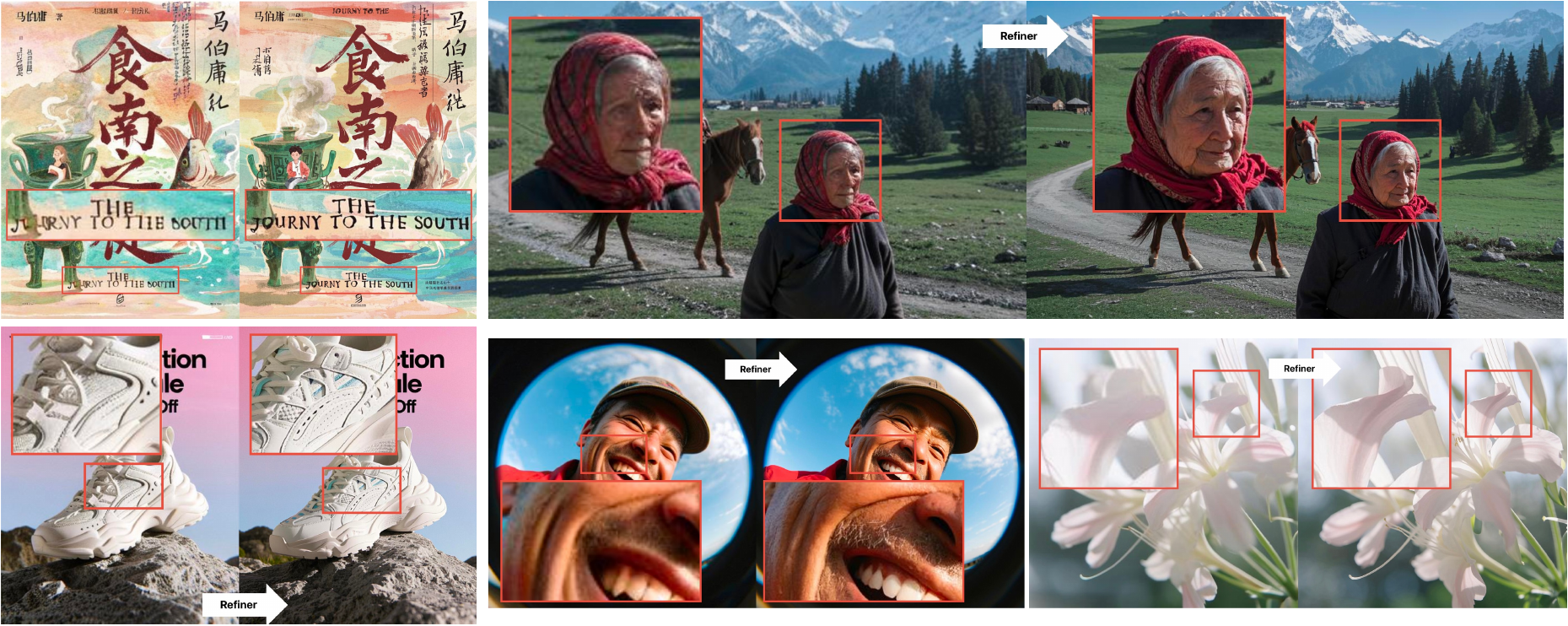}
\caption{Refiner Visualization. Recommend to zoom in for the best visualization.}
\vspace{-2pt}
\label{fig:refiner}
\end{figure*}

\subsection{Refiner}
Our base model generates a 512-resolution image, which is required to further scale up to 1024 resolution. We incorporate a refiner model to scale the images with a higher resolution. The refiner not only scales up the image resolution, but also refines structural details (such as those of human faces) and enriches the textural quality, as shown in \ref{fig:refiner}. The refiner model is built on our base model, and the training process includes two stages: 1024-resolution training and texture RLHF, which are detailed as follows.
\textbf{1024-Resolution Training.} 
We perform 1024-resolution training with data used in the CT stage, in which we exclude images with a resolution lower than 1024, while resizing higher-resolution images to 1024 by keeping their aspect ratios.
%
\textbf{Refiner RLHF.}
Further, we perform a similar RLHF process on our refiner, in an effort to enhance the textural details in the generated images. The data is constructed as follows.
We manually collected a set of high-texture images, where random degradation is performed to construct the paired data for training. Then we train a score-based texture reward model (RM) using these degraded images, and the texture RM is 
utilized to guide the optimization of a refiner model toward a richer and more meaningful generation of an image.

%% file: sections/ImageEditing.tex
\section{Align to Instruction-Based Image Editing} 

Earlier research(\cite{cao_2023_masactrl}) has shown that a Text-to-Image model excels not only in generating images, but also in comprehending images due to its inherited text-conditioned capabilities. 
Consequently, we can adapt the diffusion model into an instruction-based image editing model, further revealing its potential to benefit users.

\subsection{Preliminaries}
As introduced in SeedEdit (\cite{seededit2024}), we proposed a new data generation process, a novel causal diffusion framework, and a training strategy with iterative optimization. 
Specifically, for data generation, we propose a strategy similar to InstructPix2Pix (\cite{brooks2023instructpix2pix}) or Jedi(\cite{zeng2024jedi}), which helps ensure broader data variability including rigid and non-rigid modifications. 
In terms of architecture, we utilize the diffusion diffusion model as an image encoder, which is different from commonly used encoders such as CLIP (\cite{chen2022altclip}) or DINO (\cite{oquab2023dinov2}) as seen in methods such as IP-Adaptor (\cite{ye2023ip}). This is because we want generation features and image understanding features are aligned in the same latent domain. 

\begin{figure*}[h]
\centering
\includegraphics[width=\linewidth]{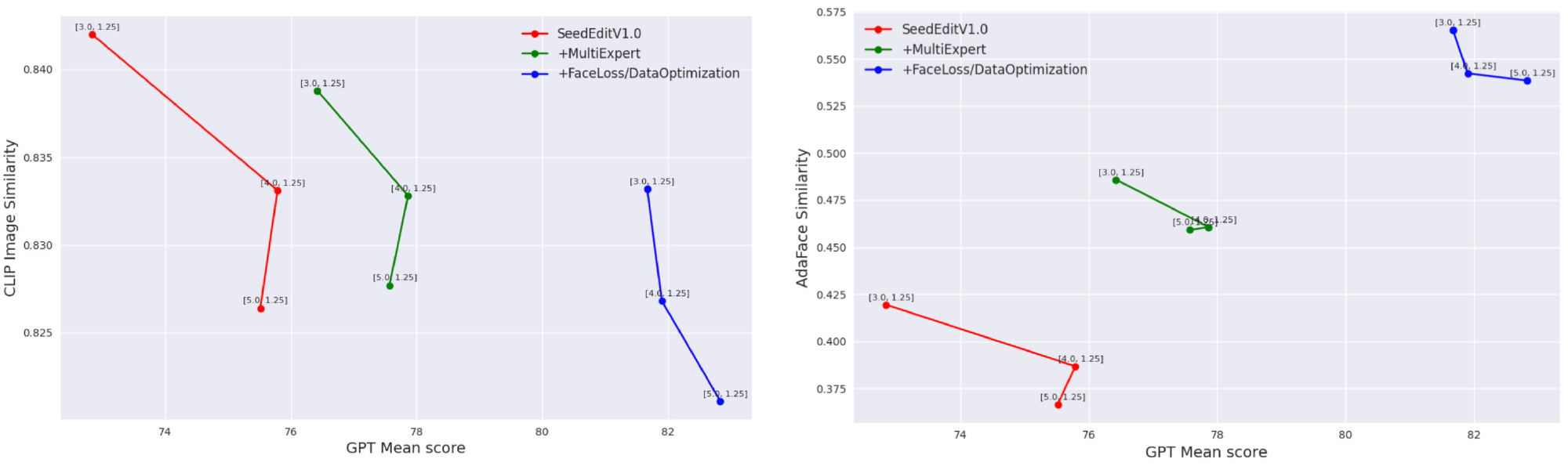} 
\caption{Quantitative ablation of SeedEdit. Left: GPT score v.s. CLIP image similarity. Right: GPT score v.s. AdaFace similarity.}
\label{fig:edit_ablation}
\end{figure*}

SeedEdit results in edited images that retain high aesthetic and compositional fidelity with the original input. Finally, we employ an iterative optimization strategy to better integrate image and textual features for generating new images. By fusing these techniques, SeedEdit delivers superior editing quality for both synthesized and real images, surpassing other state-of-the-art academic and product benchmarks. The method we outline in this paper is referred to as SeedEditV1.0, and here we have subsequent improvements detailed in this technical report.

\subsection{Enhanced Human ID Preservation}
After the launch of SeedEditV1.0, we observed limited performance in retaining human facial ID in real images, particularly when the face is small or impacted by the diffusion model's strong text-conditioned bias. 
For example, positioning a person in front of “The Taj Mahal” might drive their appearance close to an Indian face. Since human facial features are crucial for our applications, we introduce two enhancements to address this issue.

\begin{figure*}[tp]
\centering
\includegraphics[width=\linewidth]{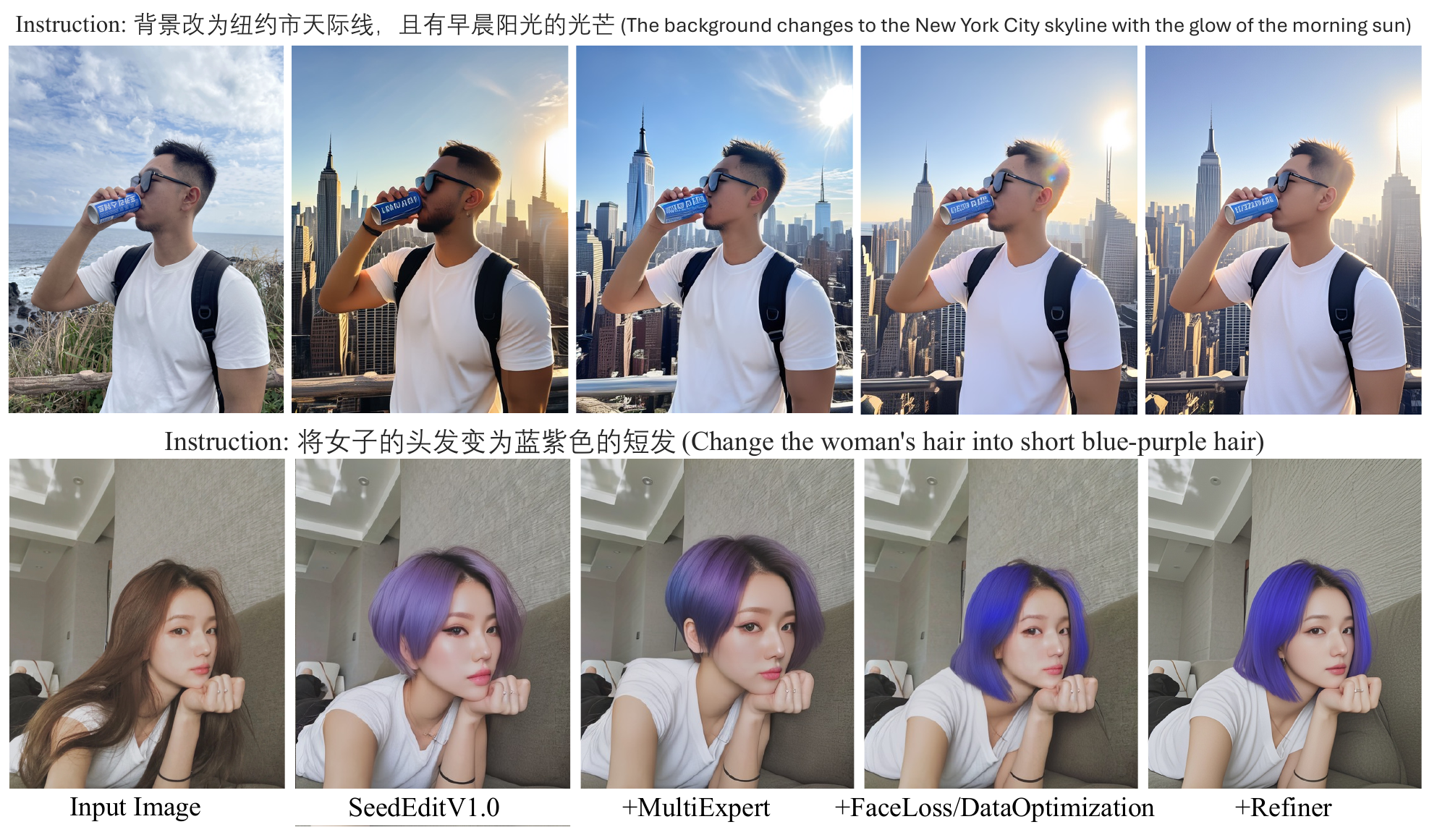} 
\caption{Qualitative comparison of SeedEdit revision. We show here that current approach significantly enhances ID retention.}
\label{fig:edit_example}
\vspace{-2pt}
\end{figure*}

\paragraph{Multi-Expert Data Fusion.} Given that generated data often includes unrealistic variations of the generated IDs, we compiled additional datasets containing real IDs from two sources. First, we created datasets using internal face-expert workflows like ID/IP guided models and background replacement models. Second, we amassed a significant dataset of real images preserving IDs, where individuals are pictured in varying environments and camera settings. 
During training, these datasets are conditionally merged based on specific data prompt prefixes to ensure the original data quality and distributions remain unaffected.

\paragraph{Face-Aware Loss.} For image pairs that preserve human face IDs well, we further boost the model's capability to maintain facial features by implementing an additional perception loss through a face similarity measurement model such as AdaFace(\cite{kim2022adaface}). By combining diffusion loss with face loss, the updated SeedEdit model markedly improves facial similarity. 

\paragraph{Data Optimization.} Lastly, we further refine the quality of the data employing more robust data filters and a wider variety of sampling strategies, resulting in an improved edit model. In our experiments, we cultivated a 160 image edit validation set with both real and generated images covering various editing operations. Figure \ref{fig:edit_ablation} illustrates the impact of expert data and face-aware loss on the SeedEdit revision, where each component improves significantly demonstrating how both strategies enhance the outcomes. Examples are listed in Figure\ref{fig:edit_example}.

%% file: sections/Model_Acceleration.tex
\section{Model Acceleration}

\subsection{CFG and Step Distillation} \label{sec: CFG_PCD}
In the diffusion model inference stage, the Classifier-Free Guidance (CFG) strategy is commonly employed, necessitating two model inferences per timestep to generate an image. To address this inefficiency while maintaining guidance scale parameterization, we propose a novel guidance scale embedding strategy. 
Our step distillation framework builds upon Hyper-SD \cite{ren2025hyper}, which introduces a novel Trajectory Segmented Consistency Distillation (TSCD) methodology for efficient diffusion model compression. TSCD employs a hierarchical refinement strategy combining trajectory preservation and reformulation mechanisms through three sequential operational phases: First, Hyper-SD divides the full timestep range $[0, T]$ into $k$ segments (initially $k=16$) for localized consistency learning, ensuring that each segment maintains the original ODE trajectory characteristics through boundary-aware temporal sampling. Then, we gradually reduces the segment count ($k\rightarrow[8,4,2,1]$) across training stages, enabling a smooth transition from local to global consistency. This hierarchical refinement mitigates error accumulation, a common issue in single-stage consistency distillation methods. Furthermore, we adaptively balance MSE for proximal predictions and adversarial loss for divergent targets. Experiments confirm improved stability and efficiency. Integrating these phases, TSCD enhances diffusion model compression while preserving high-fidelity generation.

\subsection{Quantization} \label{sec: Quantization}
We have significantly improved computational density and reduced kernel memory access through operator fusions and fine-tuning for intensive operations. These efforts have led to performance improvements of operators ranging from 5\% to 20\%. We also support Attention and GEMM quantization and propose an adaptive hybrid quantization approach. Initially, an offline smooth method is employed to optimize distribution outliers within intra-layers. Subsequently, we implemented a search strategy for various layers, bit-width, and granularities based on sensitivity metrics. Furthermore, we proposed a lightweight Quant Training scheme. Specifically, we fine-tune the quant scale in a few hours. This approach assists the low-bit model in adapting to district activation variations and further mitigates quantization losses of hard-to-smooth sensitive layers. To achieve acceleration benefits on GPUs, we optimized various low-bit mixed-granularity quantization kernels.

%% file: sections/Model_Performance.tex
\section{Model Performance}

We conducted a comparative analysis between our model and several SOTA text-to-image models. For the performance on English prompts, we compare our model with recent commercial models including GPT-4o (\cite{openai2024gpt4ocard}), Midjourney v6.1 (\cite{mjv6}), FLUX1.1 Pro (\cite{flux2023}) and Ideogram 2.0 (\cite{ideogram}). For performance on Chinese prompts, models including GPT-4o (\cite{openai2024gpt4ocard}), Kolors 1.5 (\cite{kolors}), MiracleVision 5.0 (\cite{meitu}) and Hunyuan (Dec. 2024 \cite{hunyuan}) are compared. Both human and machine evaluation are used to provide more comprehensive studies. The results show that our model exhibits remarkable proficiency in both Chinese and English, attaining the highest score on the most perspectives, and emerging as the most widely preferred model. Further evaluations focusing on text rendering and Chinese characteristics demonstrate that our model exhibits superior performance in the generation of accurate Chinese cultural
nuances and related content, surpassing current industry competitors. The overall results are presented in Figure 1.

\subsection{Human Evaluation}
\subsubsection{Benchmark}

For a comprehensive assessment of the performance of text-to-image models, a rigorous evaluation benchmark is established. This benchmark, named Bench-240, is made up of 240 prompts. These prompts are collected by combining representative prompts from publicly accessible benchmarks, such as \cite{yu2022scalingautoregressivemodelscontentrich}, and manually curated prompts. Each prompt is provided in both Chinese and English.
The design of this benchmark focuses on two considerations: image content such as subject and their relations or relevant actions, and image quality such as subject structure and aesthetic elements. 
The distribution of text prompts is meticulously calibrated in accordance with user preference surveys.

\subsubsection{Human Evaluation Results}

\begin{figure*}[h]
\centering
\includegraphics[width=\linewidth]{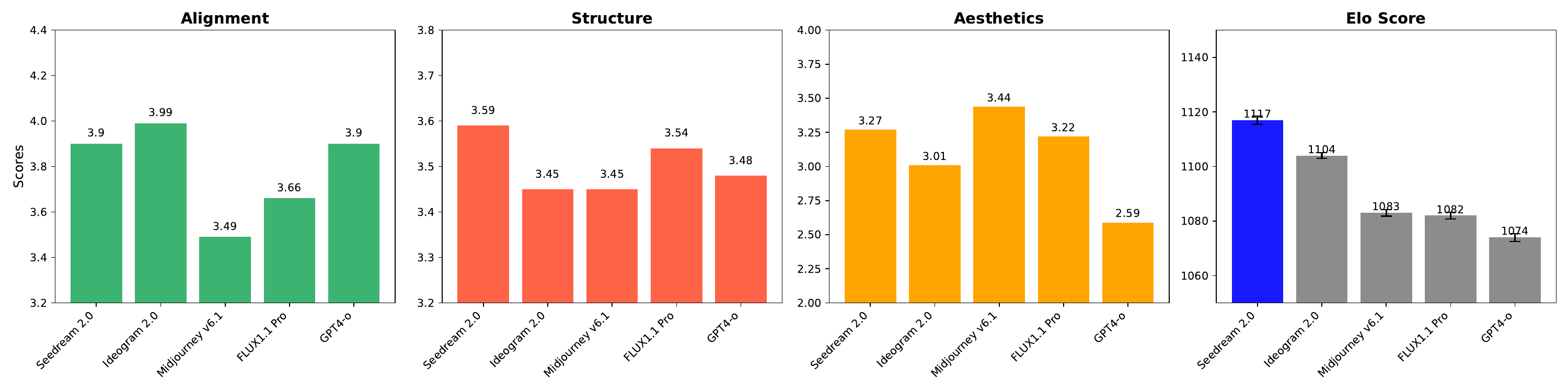}

\caption{Human Evaluation Results.}
\label{fig:human_eval_res}
\end{figure*}

Based on Bench-240, a comprehensive comparison of various models is performed by computing an overall ELO score with a professional evaluation on three key aspects: text-image alignment, structural correction, and aesthetic quality.
We report the result in Figure~\ref{fig:human_eval_res}.

\begin{itemize}[leftmargin=*]
\item ~\textbf{\textit{Expert Evaluation on specific aspects}}:
Professional evaluation is conducted by expert reviewers, who are professionals equipped with specialized skills or extensive hands-on experience in their respective domains. For example, in terms of aesthetic quality, proficient aesthetic designers are required to assign an aesthetic score to each generated image. These reviewers use a Likert scale (\cite{likert-scale}) that spans from 1 (denoting extreme dissatisfaction) to 5 (signifying utmost satisfaction) to quantitatively evaluate the generated images. The ultimate score for each model is computed as the arithmetic mean of the scores provided by multiple reviewers, across a series of images and corresponding prompts.
\item ~\textbf{\textit{Elo-based total score}}:
The overall public preference is gauged through an Elo-based (\cite{elo-ranking}) ranking system, which is calculated from the voting results of the public reviewers. Volunteers are presented with pairwise comparisons of images produced by two different models and are asked to make their preferred selection.
We have collected more than 500,000 pairwise comparisons, with each model participating in an average of more than 30,000 comparisons. It should be noted that some subversions or competing models are also involved. The extensive results provide a reliable reference to gauge public preference.
\end{itemize}

As illustrated in Figure 1, our Seedream achieves the preeminent total score among public reviewers for both Chinese and English evaluation, remaining in a considerable superiority over the remaining models.
Furthermore, the Seedream model exhibits a more holistic performance across all evaluation rubrics. 
Taking the English evaluation as an example, we present more specific results in Figure ~\ref{fig:human_eval_res}.
Our model ranks first in structural aspects and occupies the second position in both image-text alignment and aesthetic performance. It has no obvious shortcomings, which excells over Midjourney v6.1 in image-text alignment and Ideogram 2.0 in aesthetic appeal. While competing models possess fortes in particular dimensions, our model preponderates across the entire spectrum of criteria, thereby emerging as the most favored in public appraisals. Similar conclusion can be seen in the Chinese evaluation as well. Some comparison images are shown in Figure \ref{fig:Alignment_Comparisons},\ref{fig:Structure_comparisons},\ref{fig:Aesthetics_comparisons}.

\subsection{Automatic Evaluation}
Automated evaluation techniques are additionally utilized to assess the performance of text-to-image models, especially those that are publicly available. Our evaluation principally takes into account two aspects: text-image alignment and image quality. Only English prompt results are presented here, since external automatic evaluation methods mainly support English input.

\begin{figure*}[h]
\centering
\includegraphics[height=8cm]{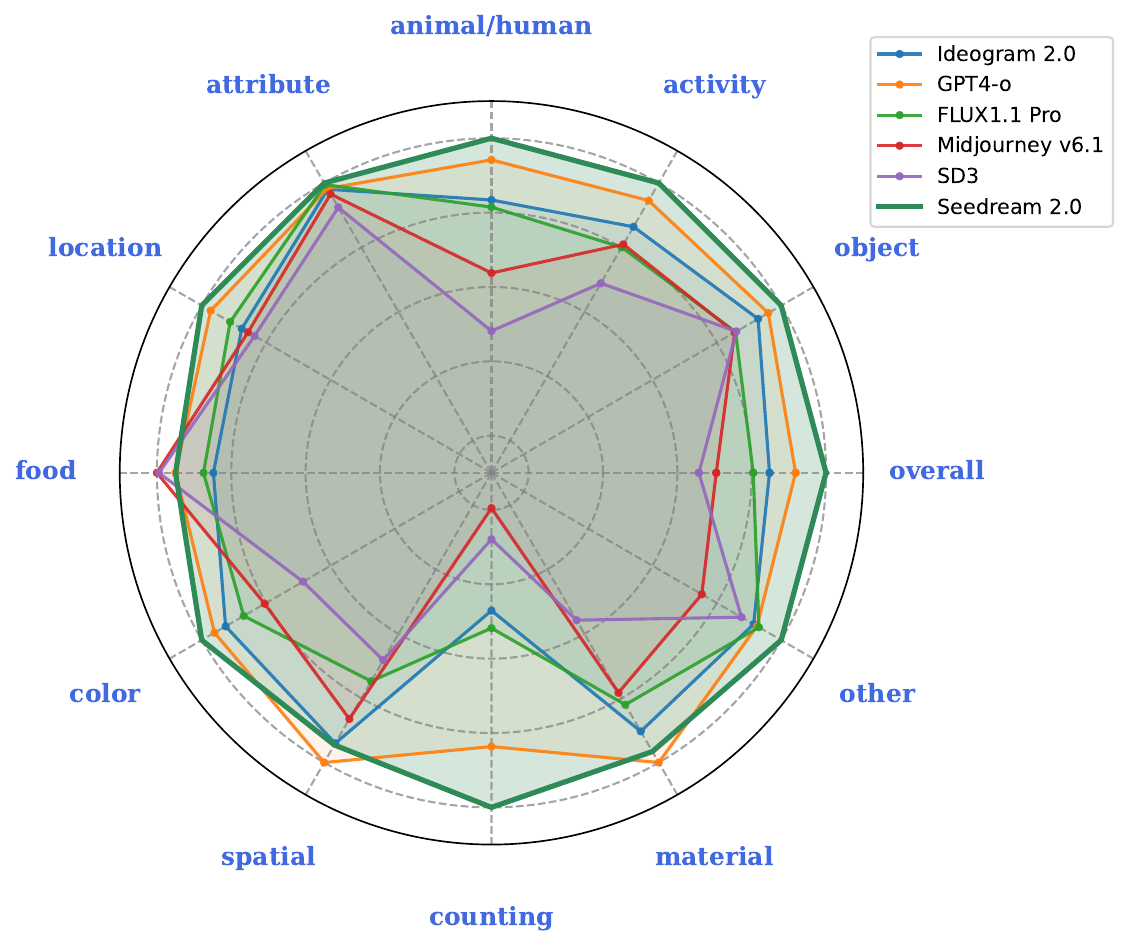}
\caption{EvalMuse Evaluation Results across fine-grained dimensions.}
\label{fig:evalmuse_radar}
\end{figure*}

\subsubsection{Text-Image Alignment}
Traditional metrics like FID (\cite{jayasumana2024rethinkingfidbetterevaluation}) and CLIP-Score (\cite{hessel2022clipscorereferencefreeevaluationmetric}) prove inadequate in precisely measuring the image-text alignment capabilities of current text-to-image models. Consequently, automated evaluation methodologies harnessing Vision Language Models (VLMs) have attracted substantial interest. In the present study, we adopt two approaches: EvalMuse (\cite{han2024evalmuse40kreliablefinegrainedbenchmark}) and VQAScore (\cite{lin2024evaluating}).
\begin{itemize}[leftmargin=*]
\item ~\underline{\textit{EvalMuse}}:
EvalMuse collects and annotates an extensive dataset of image-text pairs, facilitating a detailed analysis of image-text alignment within generated images. By employing the FGA-BLIP2 model, which exhibits a remarkable degree of consistency with human evaluations across multiple benchmarks, we conduct a comparison of diverse models on the EvalMuse test dataset and present fine-grained results among skill dimensions.
\item ~\underline{\textit{VQAScore}}:
VQAScore capitalizes on a visual-question-answering (VQA) model to derive alignment scores by computing the probability of whether the generated image corresponds to the prompt. Driven by state-of-the-art Vision-Language Models (VLMs), VQAScore attains an accuracy level comparable to that of human evaluations. In this study, we utilize the recommended Clip-Flant5-xxl model to automatically evaluate the image-text alignment capabilities on GEN-AI benchmark.
\end{itemize}

\setlength{\tabcolsep}{1pt}
\begin{table*}[htbp]
  \centering\textit{}
    \begin{tabular}{m{3cm}|m{2cm}|m{1.2cm}|cccccccc}
    \toprule
     & \centering VQAScore & &&&& EvalMuse &&& \\
   \centering Model & \centering total & \centering Total & Object  & Activity & a./h. & Attribute & Location  & Color & Counting & Other \\
    \midrule
 \centering Seedream 2.0 & \centering \centering \underline{0.8031} & \centering \textbf{0.682} & \textbf{0.747} & \textbf{0.662} & \textbf{0.756} & \textbf{0.821} & \textbf{0.793} & \textbf{0.706}& \textbf{0.477} & \textbf{0.665} \\
    
 \centering GPT-4o & \centering 0.7974 & \centering 0.656 & 0.732 & 0.644 & 0.734  & 0.814  & 0.782 & 0.692 & 0.438  & 0.640  \\

 \centering FLUX1.1 Pro & \centering 0.7877 &  \centering 0.617 &  0.694 & 0.596 & 0.686 & 0.819 & 0.758 & 0.660 & 0.362 & 0.642 \\

 \centering Ideogram 2.0 & \centering \textbf{0.8226} & \centering 0.632 & 0.720 & 0.617 & 0.693 & 0.813 & 0.743 & 0.680 & 0.351 & 0.637\\
 
 \centering Midjourney v6.1 & \centering 0.7569 & \centering 0.583  & 0.693  & 0.599  & 0.619 & 0.807  & 0.736 & 0.637 & 0.285 & 0.583 \\
    \bottomrule
    \end{tabular}%
      \caption{Automatic evaluation results using VQAScore and EvalMuse.
  }
  \label{tab:t2i}%
\end{table*}%

VQAScore gives similar results to those of human evaluations, that our Seedream ranks second only to Ideogram and is ahead of other models. The findings from the EvalMuse assessment reveal that our model achieves the highest composite score, claiming the top position across the majority of crucial metrics. 
Especially in some dimensions with higher difficulty, such as counting and activity. In addition, our model also takes the lead over other models in the "other" category, because the text rendering ability is also included in this category.
Significantly, the results from automated evaluations closely mirror manual appraisals, further validating that our model has excellent performance in handling image-text alignment.

\subsubsection{Image Quality}
Image quality is highly subjective, thereby posing a significant challenge in formulating a universally applicable and accurate standard for evaluation. Conventionally, human preference metrics have been resorted to for assess the visual appeal of an image. In this study, we evaluate the performance of our model by the following models: HPSv2 (\cite{wu2023human}) and MPS (\cite{MPS}).
\begin{itemize}[leftmargin=*]
\item ~\underline{\textit{HPSv2}}: derives from an expansive dataset of annotated generated image pairs, it proffers a steady and dependable measure of image quality.
\item ~\underline{\textit{MPS}}: Conversely, this metric evaluates image quality across multiple dimensions, and it has been demonstrated that it exhibits especially potent discriminatory capabilities in capturing aesthetic perception.
\item ~\underline{\textit{Internal Evaluation Model}}: Additionally, we introduce two internally preferred evaluation models, namely Internal-Align and Internal-Aes, which are respectively utilized for the evaluation of text-image alignment and overall aesthetic aspects.
\end{itemize}

\begin{table*}[]
  \centering
\begin{tabular}{p{2.5cm}|p{2cm} p{2cm} p{2.5cm} p{2cm} p{2cm} p{2cm}}
    \toprule
    Metirc & ~GPT-4o & FLUX 1.1  & Ideogram 2.0 & MJ v6.1 & RecraftV3 & Seedream 2.0 \\
    \midrule
 HPSv2 & ~~0.2881&~0.2946&~~0.2932&~0.2850&~0.2991&~~\textbf{0.2994} \\
 MPS & ~~12.79 &~13.11&~~13.01&~\textbf{13.67}&~13.09&~~\underline{13.61}  \\
 \midrule
 Internal-Align & ~~28.85 &~27.75&~~27.92&~\underline{28.93}&~28.90&~~\textbf{29.05}  \\
Internal-Aes & ~~26.48 &~25.15&~~26.40&~\textbf{27.07}&~26.80&~~\underline{26.97}  \\

    \bottomrule
    \end{tabular}
      \caption{Preference Evaluation with different metrics.}
  \label{tab:image_quality}%
\end{table*}%

We present the quality evaluation results for these two metrics on Bench-240, comparing our model against GPT-4o, FLUX-1.1, Midjourney v6.1 and RecraftV3. Our model achieves the highest score on HPSv2. In terms of the MPS score, our model trails closely behind Midjourney v6.1, yet outperforms other competing models by a substantial margin. 
Similar trends can be observed in the internal evaluation model. Whereas the performance of competing models exhibits significant oscillations across diverse evaluation metrics, our model demonstrates remarkable stability and persistently high performance, underlining its preponderant capability across a spectrum of preference dimensions.

\subsection{Text Rendering}
To comprehensively assess the text rendering capability of our model, we carried out an extensive evaluation as well. Initially, we devised a specialized benchmark tailored for text rendering, which incorporated 180 prompts in Chinese and an equal number in English. These prompts encompass a wide range of categories, ranging from logo designs, posters, electronic displays, printed text, to handwritten text. Notably, the benchmark also contains text renderings on unconventional substrates, such as text formed by arranging French fries or inscribed in the semblance of clouds, thereby providing a diverse and comprehensive benchmark.

One subjective metric, \textbf{availability rate}, and two objective metrics, text \textbf{accuracy rate} and \textbf{hit rate}, are employed to evaluation of text rendering capability.
Availability rate refers to the proportion of images deemed acceptable when text rendering is almost correct, taking into account the integration of text with other content and the overall aesthetic quality. The objective metrics are defined as follows:

\begin{itemize}
\vspace{-1mm}
    \item \textbf{Text accuracy rate} is defined as:
    \vspace{-1mm}
    \[
    R_a = (1 - \frac{N_e}{N}) * 100\%
    \]

    where \(N\) represents the total number of target characters, and \(N_e\) denotes the minimum edit distance between the rendered text and the target text.
    \item \textbf{Text hit rate} is defined as:
    \[
    R_h = \frac{N_c}{N} * 100\%
    \]
    \vspace{-1mm}
    where \(N_c\) represents the number of characters correctly rendered in the output.
    \vspace{-1mm}
\end{itemize}

To deeply assess our model's text rendering capabilities, we carefully compared it with outstanding text-to-image models having text rendering features. In English text rendering, the competitors included RecraftV3 (\cite{recraftv3}), Ideogram 2.0, FLUX1.1 Pro, GPT-4o, and Midjourney v6.1. For Chinese aspects, the evaluation covered Kolors 1.5 (\cite{kolors}) and MiracleVision 5.0 ({\cite{meitu}}).

\begin{figure*}[h]
\centering
\includegraphics[width=\linewidth]{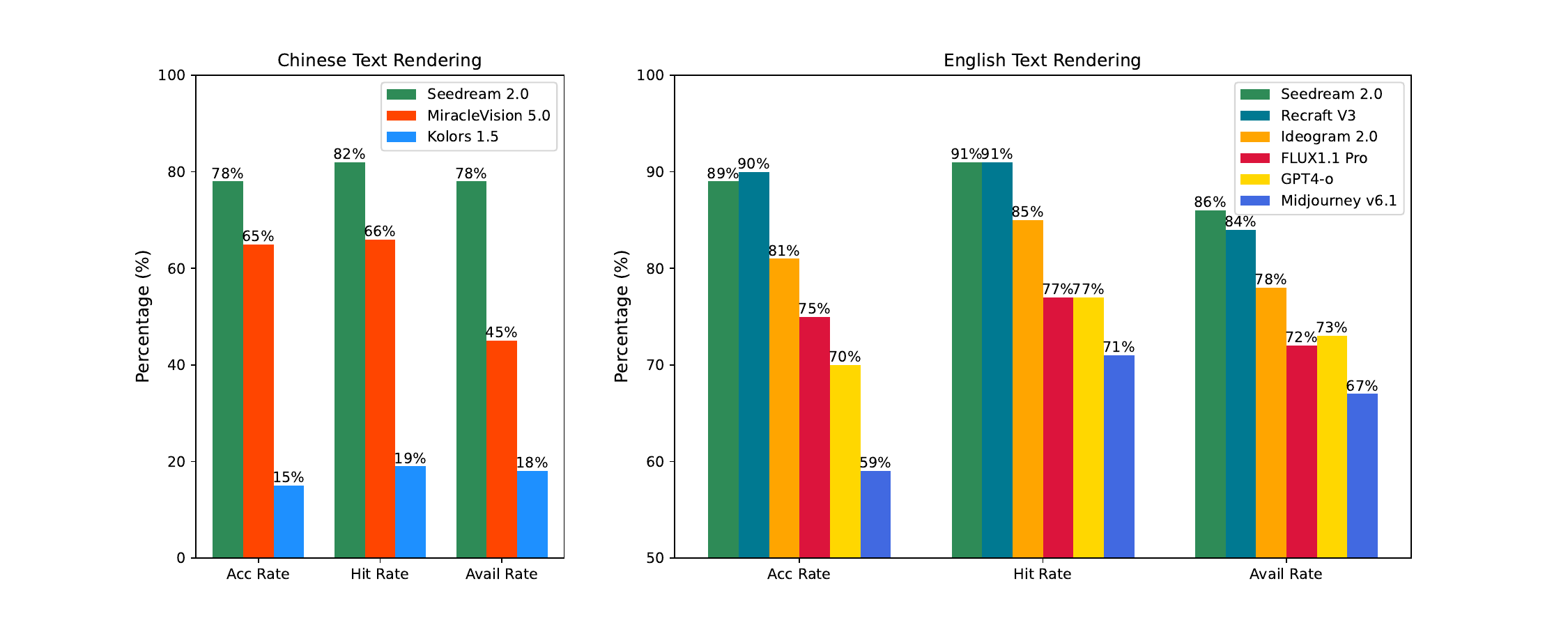}
\caption{Text Rendering Evaluation.}
\label{fig:text_rendering}
\end{figure*}

The evaluation results, Figure~\ref{fig:text_rendering}，clearly show that our model achieves the best availability in both Chinese and English text rendering, with the highest or near-highest text accuracy and hit rates among the tested models. Especially in Chinese text rendering, our model gives a clear edge over all rivals.
Compared to generating English characters, rendering Chinese characters is much more challenging due to their more complex structures and a much larger character set. Despite these difficulties, our model achieves an impressive 78\% text accuracy rate and 82\% hit rate in Chinese writing.
Although MiracleVision 5.0 also achieves a 65\% Chinese text accuracy rate, its text layout often keeps an obvious disconnect from the background of the image, seriously affecting availability.
Moreover, our model stands out by being great at generating Chinese text with rich cultural meanings, like traditional couplets and ancient Chinese poetry, highlighting its ability to handle special and nuanced text forms. Examples can be found in Figure~\ref{fig:text_rendering_comp} and Figure~\ref{fig:text_rendering_show}.

\subsection{Chinese Characteristics}
Generating images that accurately describe Chinese characteristics requires not only a basic understanding of the Chinese language, but also a nuanced perception of China's rich cultural heritage. For example, ancient China cannot be represented by a single symbol, as each dynasty (Tang, Song, Yuan, Ming, and Qing) has distinct cultural features.
To comprehensively evaluate the performance of our model in terms of Chinese characteristics,  we construct a benchmark of 350 prompts that span traditional clothing, food, artistic techniques, architecture, and other customs. 

\begin{figure*}[h]
\centering
\includegraphics[height=6cm]{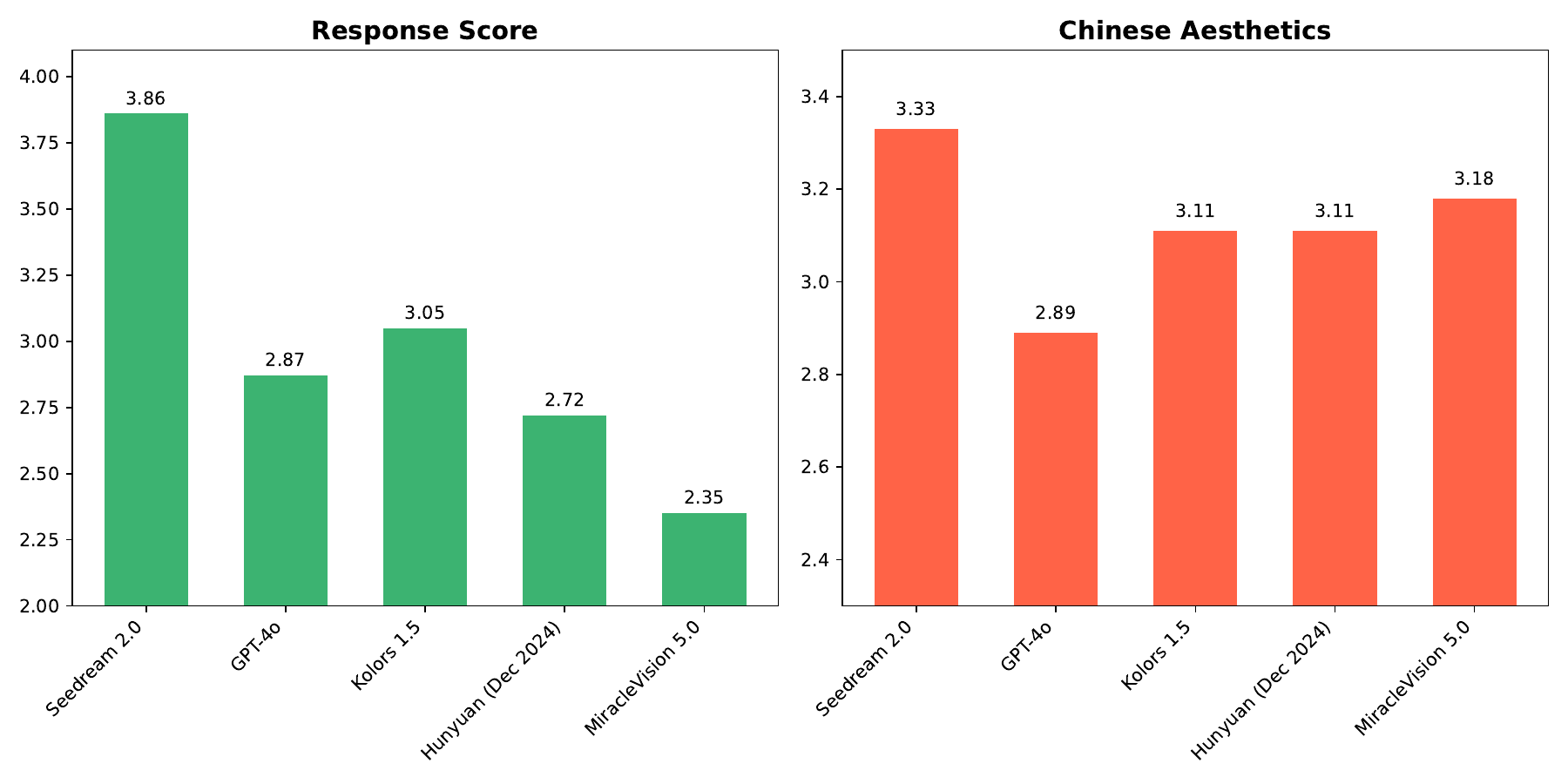}
\caption{Chinese Characteristics Evaluation.}
\label{fig:chinese_cha}
\end{figure*}

Professional designers evaluate the generated images based on two criteria. A response rate indicates whether the target elements are correctly responsed to. A Chinese aesthetic score refers to whether the expression by the generated images satisfies the aesthetic tendencies in China. Both metrics are scored on a scale from 1 to 5, with 1 representing no response and 5 indicating a perfect meeting.

\begin{figure*}[h]
\centering
\includegraphics[height=8cm]{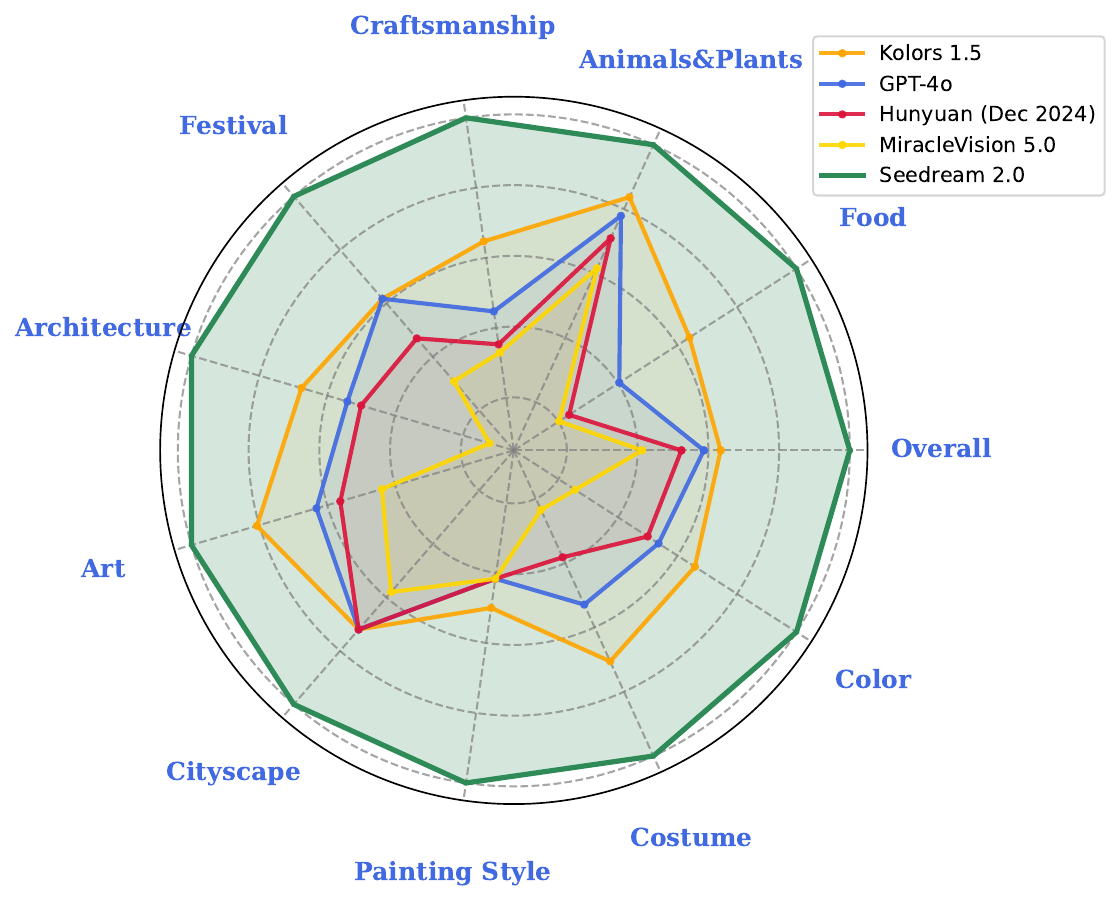}
\caption{Response Rate of Chinese Characteristics across Dimensions.}
\label{fig:chinese_dimension}
\end{figure*}

Figure~\ref{fig:chinese_cha} shows that our model outperforms the others, particularly in response rate, with a clear advantage.
We further analyze the proportion of correct responses (with a response score of 5) of each model in a fine-grained perspective of Chinese characteristics, and the results are shown as a normalized radar~\ref{fig:chinese_dimension}. Our model significantly outperforms competitors in all dimensions, especially in aspects such as food, festival, craftsmanship, and architecture.
As shown in Figure~\ref{fig:chinese_cha_comp}, take Hot Dry noodles vs. Sliced noodles, and Mongolian vs. Tibetan robes. Other models struggle to show such differences. More high-aesthetic Chinese-style images generated by Seedream can be found in Figure~\ref{fig:chinese_char_show}.

\subsection{Visualization}
\begin{figure*}[t]
\centering
\includegraphics[width=\linewidth]{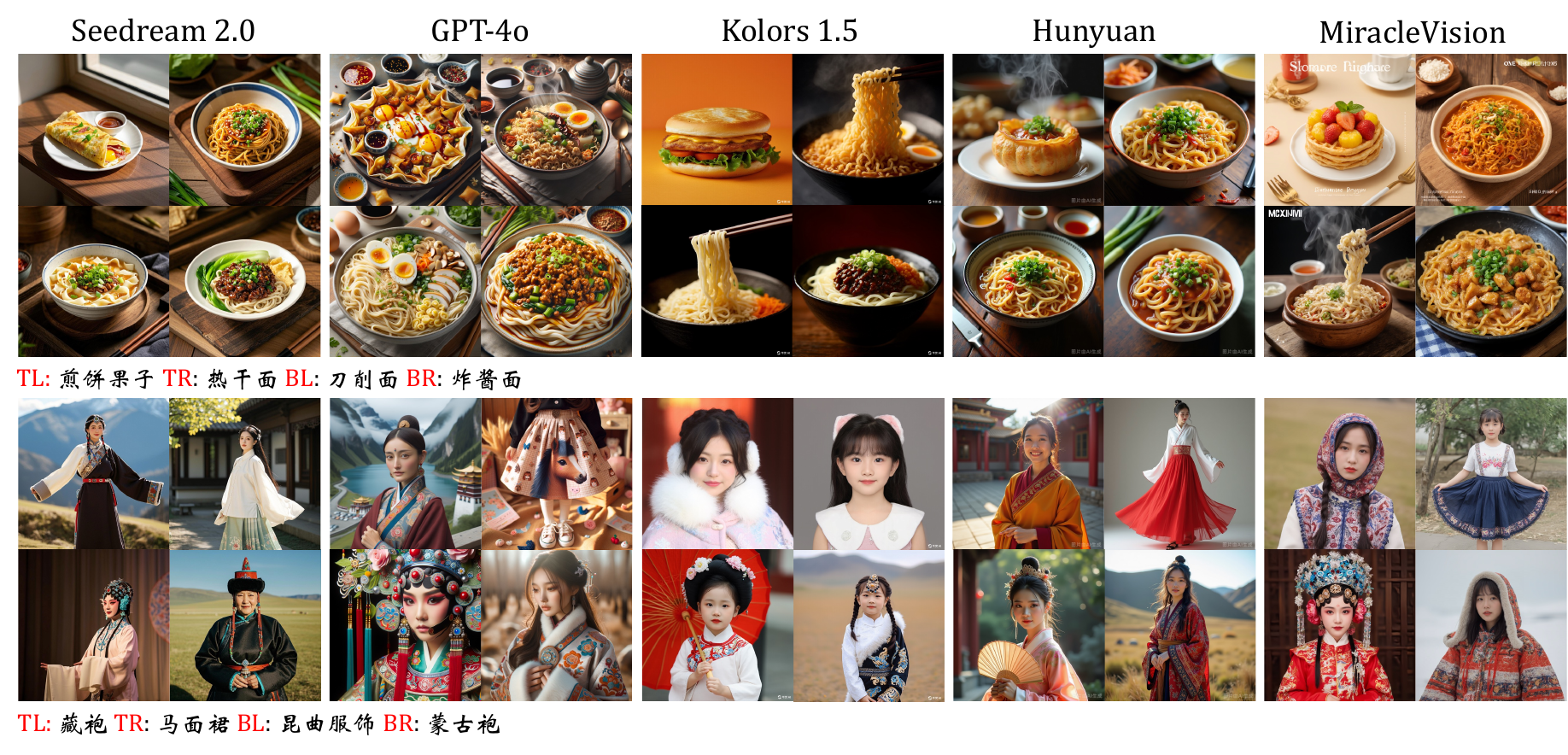}
\caption{Chinese Characteristics Comparisons. Our model demonstrates a more accurate understanding and expression of Chinese elements.}
\label{fig:chinese_cha_comp}
\end{figure*}

\begin{figure*}[t]
\centering
\includegraphics[width=\textwidth]{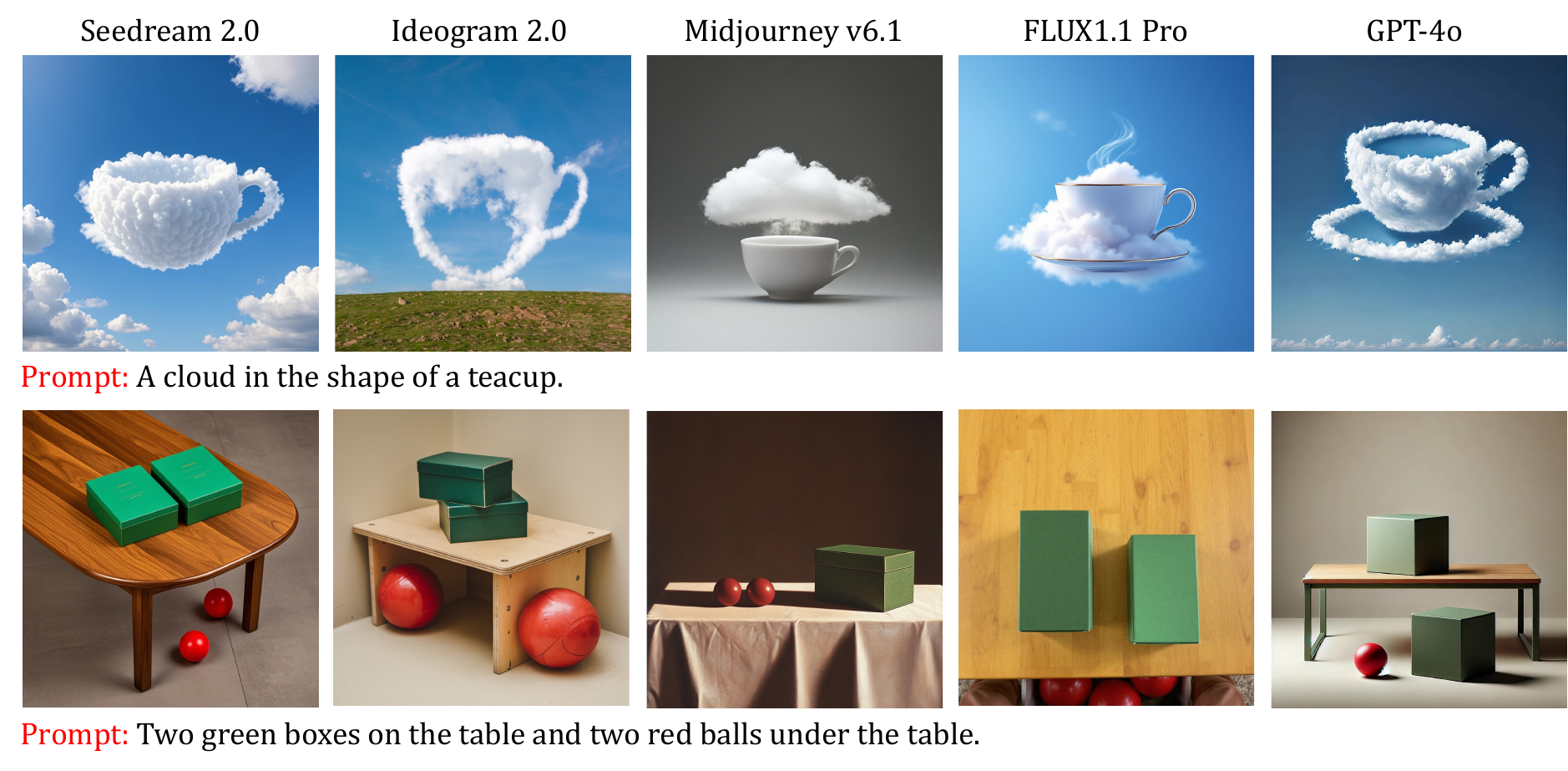}
\caption{Alignment Comparisons. Seedream and Ideogram 2.0 excel in these two prompts, while other models either struggle with imaginative scenarios or misinterpret quantity and position in the prompts below.}
\label{fig:Alignment_Comparisons}
\end{figure*}

\begin{figure*}[t]
\centering
\includegraphics[width=\textwidth]{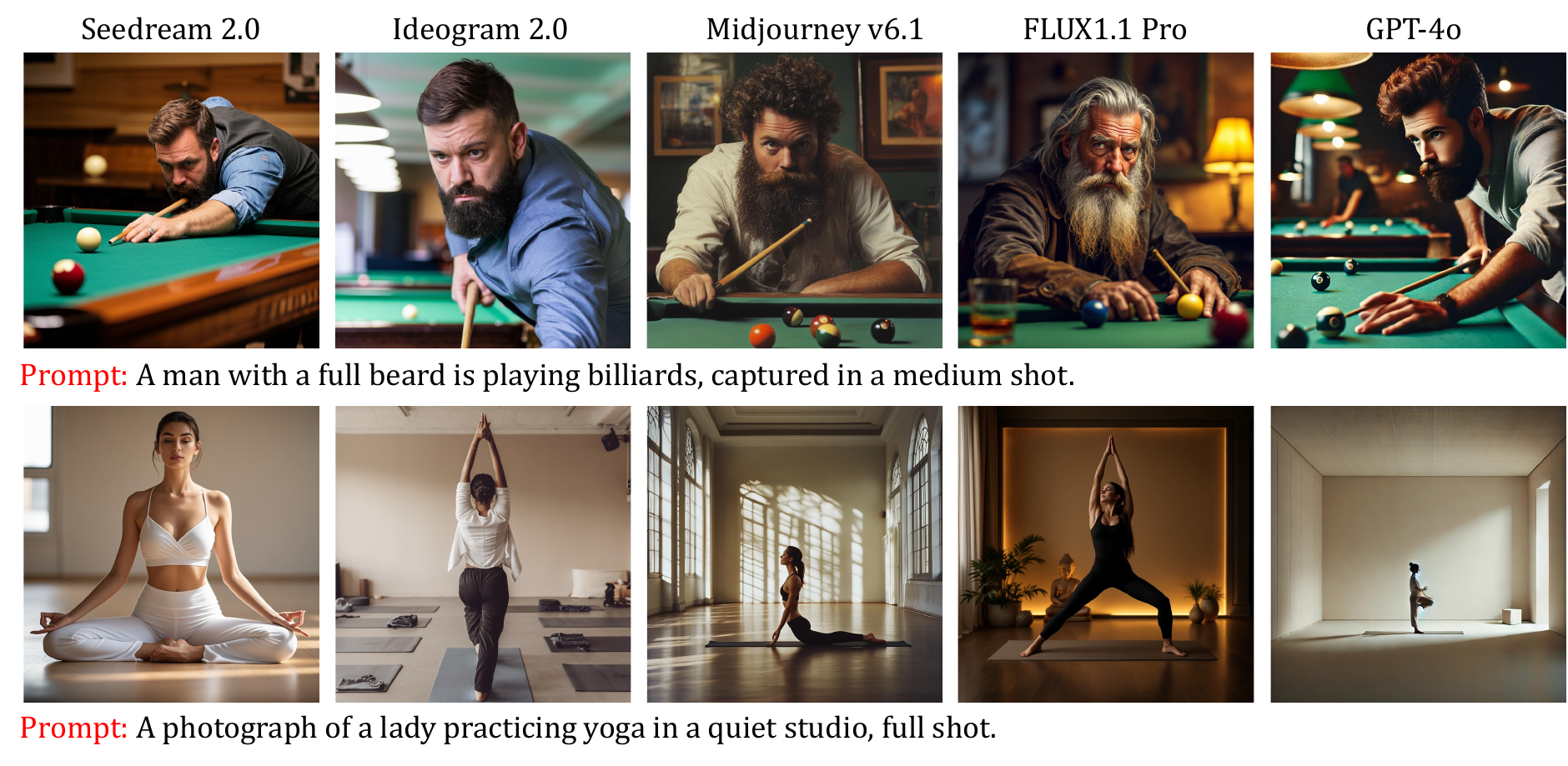}
\caption{Structure comparisons. External models encounter issues with the distortion of fingers and limbs under complex movements.}
\label{fig:Structure_comparisons}
\end{figure*}

\begin{figure*}[t]
\centering
\includegraphics[width=\textwidth]{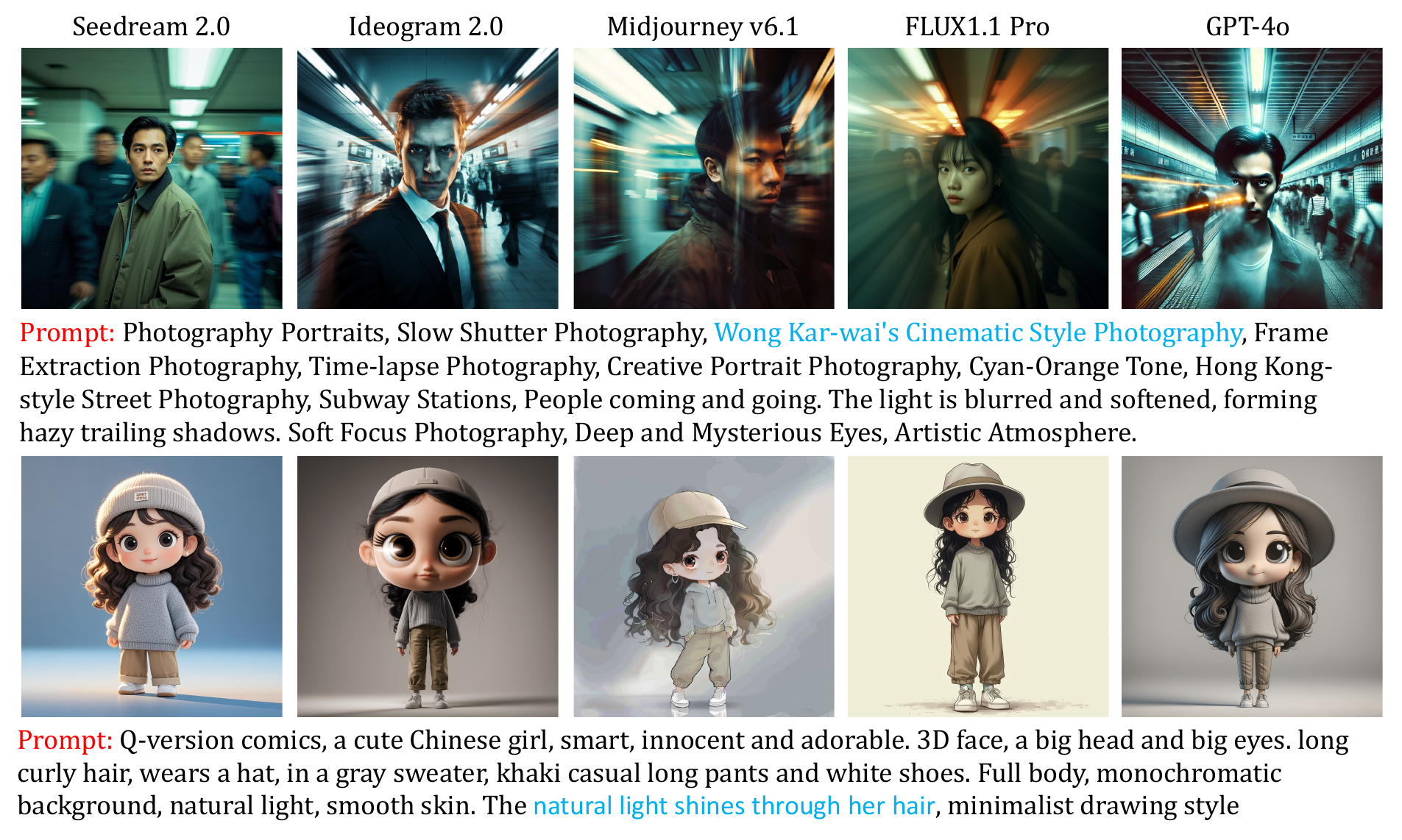}
\caption{Aesthetics comparisons. Seedream demonstrates outstanding performance in cinematic scenes and artistic design, while other models show weaker performance in artistic style and texture details.}
\label{fig:Aesthetics_comparisons}
\end{figure*}

\begin{figure*}[t]
\centering
\includegraphics[width=\linewidth]{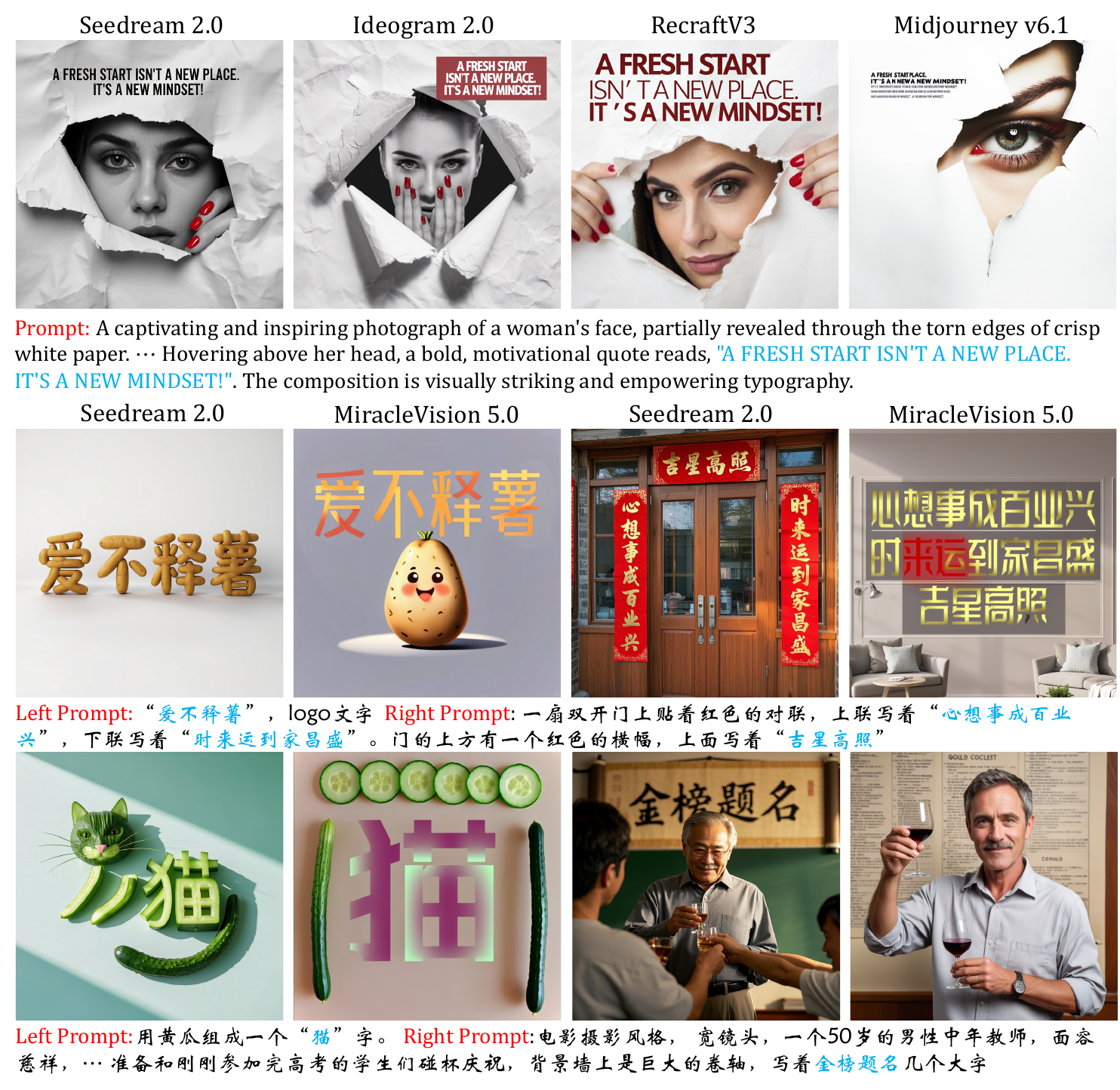}
\caption{Text-Rendering Comparisons. Seedream performs exceptionally well in harmonizing text with content and demonstrates strong typesetting capabilities. Notably, it offers a distinct understanding of scenarios with Chinese characteristics.}
\label{fig:text_rendering_comp}
\end{figure*}

\begin{figure*}[h]
\centering
\includegraphics[width=\linewidth]{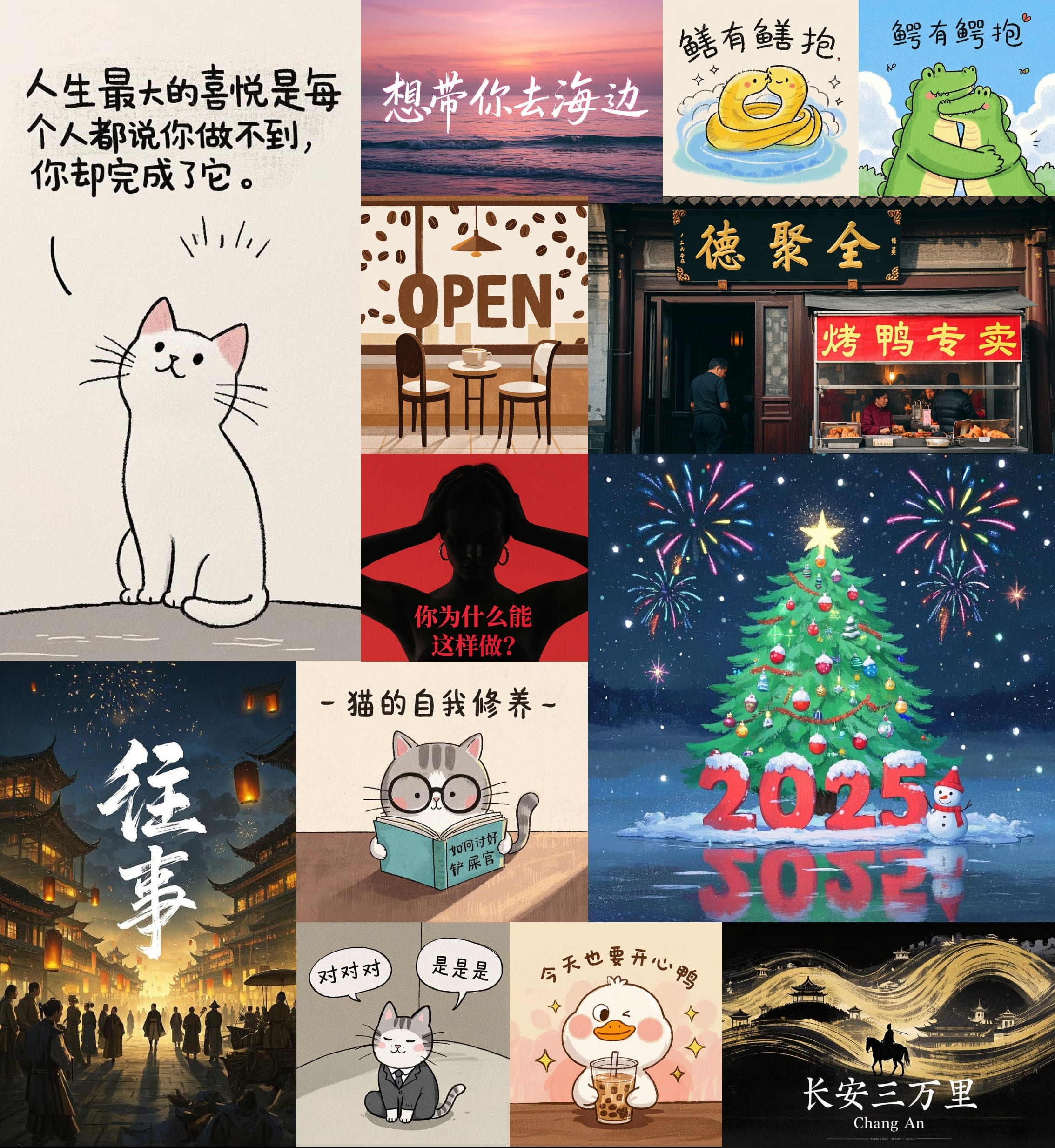}
\caption{Text Rendering by Seedream. Our model presents infinite potential in poster design and artistic creation.}
\label{fig:text_rendering_show}
\end{figure*}

\begin{figure*}[h]
\centering
\includegraphics[width=\linewidth]{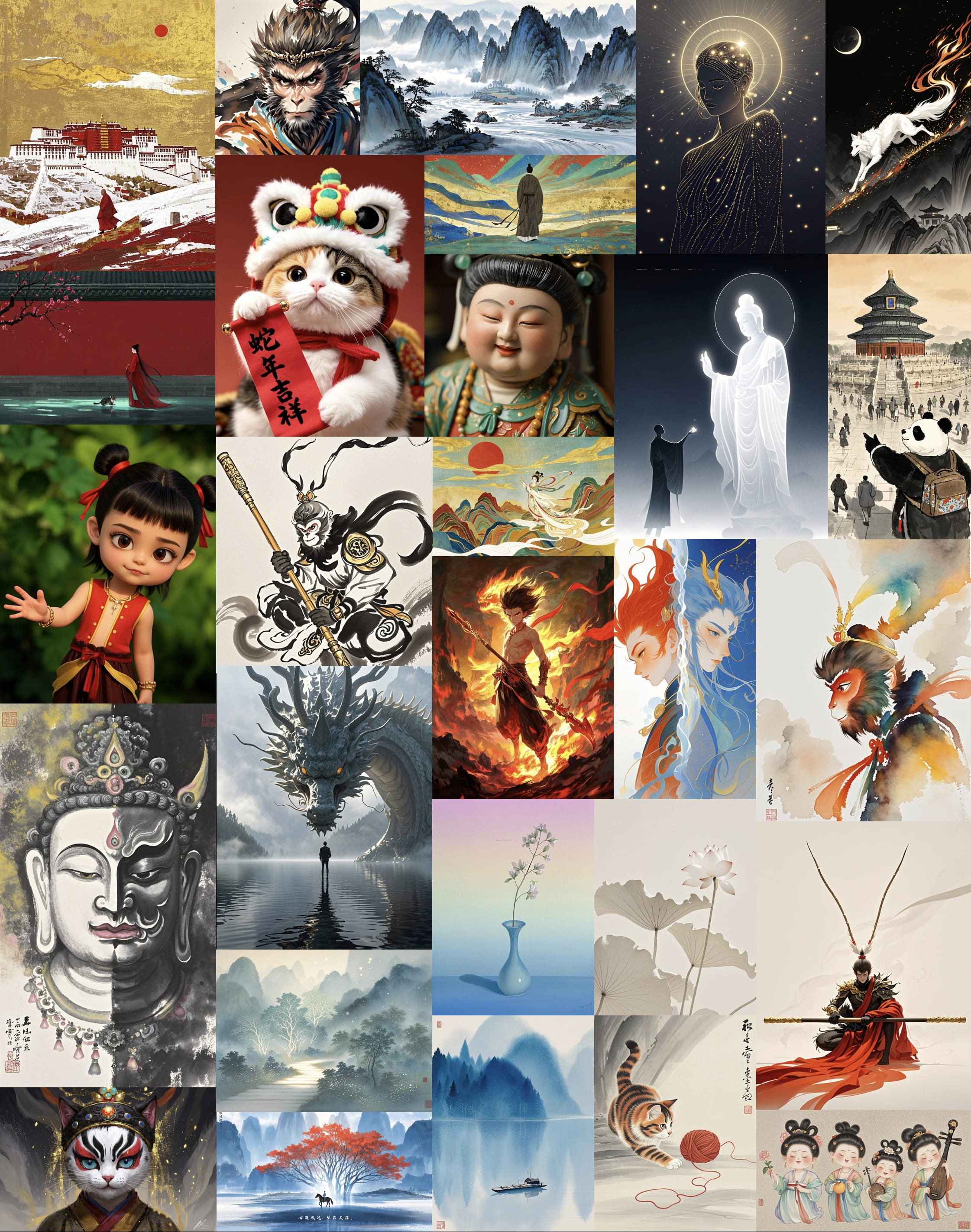}
\caption{Chinese Characteristics by Seedream. Our model presents impressive representation of Chinese aesthetics.}
\label{fig:chinese_char_show}
\end{figure*}

We present several visual comparison outcomes between our proposed approach and other existing methods in Figure \ref{fig:chinese_cha_comp},\ref{fig:Alignment_Comparisons},\ref{fig:Structure_comparisons},\ref{fig:Aesthetics_comparisons},\ref{fig:text_rendering_comp},\ref{fig:text_rendering_show},\ref{fig:chinese_char_show}. It can be seen that our approach demonstrates superiority in aspects such as image-text alignment, structural coherence, aesthetic appeal, and text rendering accuracy. For a more comprehensive exploration of our model, we invite you to visit our DouBao and Dreamina web pages.

\section{Conclusion}

In this work, we present Seedream 2.0, a state-of-the-art bilingual text-to-image diffusion model designed to address critical limitations in current image generation systems, including model bias, insufficient text rendering capabilities, and deficiencies in understanding culturally nuanced prompts. By integrating a self-developed bilingual LLM as a text encoder, our model learns meaningful native knowledge in both Chinese and English, enabling high-fidelity generation of culturally relevant content. The incorporation of Glyph-Aligned ByT5 for character-level text rendering and Scaled ROPE for resolution generalization further enhances its versatility and robustness. Through systematic optimization via multi-phase SFT and RLHF iterations, Seedream 2.0 demonstrates superior performance in prompt adherence, aesthetic quality, structural correctness, and human preference alignment, as evidenced by its exceptional ELO scores. In particular, it achieves remarkable effectiveness in Chinese text rendering and culturally specific scene generation, earning widespread acclaim on applications such as Doubao (豆包) and Dreamina (即梦).

%% file: sections/appendix.tex
\section{Contributions and Acknowledgments}
\label{contributions}

\textbf{\color{seedblue}All contributors of Seedream are listed in alphabetical order by their last names.}

\begin{multicols}{2} %
\sffamily{\color{seedblue}  \large{Core Contributors}} \\
\\
\color{seedblue}	Gong, Lixue	\\
\color{seedblue}	Hou, Xiaoxia	\\
\color{seedblue}	Li, Fanshi	\\
\color{seedblue}	Li, Liang	\\
\color{seedblue}	Lian, Xiaochen	\\
\color{seedblue}	Liu, Fei	\\
\color{seedblue}	Liu, Liyang	\\
\color{seedblue}	Liu, Wei	\\
\color{seedblue}	Lu, Wei	\\
\color{seedblue}	Shi, Yichun	\\
\color{seedblue}	Sun, Shiqi	\\
\color{seedblue}	Tian, Yu	\\
\color{seedblue}	Tian, Zhi	\\
\color{seedblue}	Wang, Peng	\\
\color{seedblue}	Wang, Xun	\\
\color{seedblue}	Wang, Ye	\\
\color{seedblue}	Wu, Guofeng	\\
\color{seedblue}	Wu, Jie	\\
\color{seedblue}	Xia, Xin	\\
\color{seedblue}	Xiao, Xuefeng	\\
\color{seedblue}	Yang, Linjie	\\
\color{seedblue}	Zhai, Zhonghua	\\
\color{seedblue}	Zhang, Xinyu	\\
\color{seedblue}	Zhang, Qi	\\
\color{seedblue}	Zhang, Yuwei	\\
\color{seedblue}	Zhao, Shijia	\\
\\
\noindent
\sffamily{\color{seedblue}  \large{Project Leader}} \\
\\
\color{seedblue}	Huang, Weilin	\\
\color{seedblue}	Yang, Jianchao	\\
\\

\noindent
\sffamily{\color{black}  \large{Contributors}} \\
\\
\color{black}	Chen, Haoshen	\\
\color{black}	Chen, Kaixi	\\
\color{black}	Dong, Xiaojing	\\
\color{black}	Fang, Jing	\\
\color{black}	Gao, Yu	\\
\color{black}	Ge, Yongde	\\
\color{black}	Guo, Chaoran	\\
\color{black}	Guo, Meng	\\
\color{black}	Guo, Qiushan	\\
\color{black}	Guo, Shucheng	\\
\color{black}	Jin, Lurui	\\
\color{black}	Kuang, Huafeng	\\
\color{black}	Li, Bo	\\
\color{black}	Li, Huixia	\\
\color{black}	Li, Jiashi	\\
\color{black}	Li, Kejie	\\
\color{black}	Li, Ying	\\
\color{black}	Li, Yiying	\\
\color{black}	Li, Yameng	\\
\color{black}	Lin, Heng	\\
\color{black}	Ling, Feng	\\
\color{black}	Liu, Shu	\\
\color{black}	Liu, ZuXi	\\
\color{black}	Lu, Hanlin	\\
\color{black}	Ou, Tongtong	\\
\color{black}	Qin, Ke'er	\\
\color{black}	Ren, Yuxi	\\
\color{black}	Wang, Rui	\\
\color{black}	Wang, Xuanda	\\
\color{black}	Wang, Yinuo	\\
\color{black}	Yao, Yao	\\
\color{black}	Zhao, Fengxuan	\\

\end{multicols} %